\documentclass{article}
\usepackage{algorithm}
\usepackage{algorithmic}

\PassOptionsToPackage{numbers, compress}{natbib}

\usepackage[preprint]{neurips_2026}

\usepackage[utf8]{inputenc}
\usepackage[T1]{fontenc}

\usepackage{amsmath}
\usepackage{amssymb}
\usepackage{amsfonts}
\usepackage{mathtools}
\usepackage{amsthm}

\usepackage{graphicx}
\usepackage{subcaption}
\usepackage{booktabs}
\usepackage{array}
\usepackage{multirow}
\usepackage{makecell}
\usepackage{adjustbox}

\usepackage{xcolor}
\usepackage{textcomp}
\usepackage{colortbl}

\usepackage{algorithm}
\usepackage{enumitem}
\usepackage{nicefrac}
\usepackage{microtype}
\usepackage{float}
\usepackage{stfloats}
\usepackage{changepage}
\usepackage{comment}
\usepackage{verbatim}
\usepackage{lipsum}

\usepackage{natbib}

\usepackage{url}
\usepackage{hyperref}
\usepackage[capitalize,noabbrev]{cleveref}

\definecolor{HeaderGray}{RGB}{245,245,245}
\definecolor{CorrectGreen}{RGB}{0,170,90}

\newcommand{\methodcolw}{0.1\textwidth}
\newcommand{\topcolw}{0.166\textwidth}

\newcommand{\imgformvspace}{-5pt} 
\newcommand{\rowgap}{2pt}         

\newcommand{\formula}[1]{{\footnotesize #1}}

\newcommand{\methodcellh}{2cm} 
\newcommand{\methodraise}{0pt}

\newcommand{\methodBox}[1]{%
  \parbox[c][\methodcellh][c]{\linewidth}{%
    \centering
    \raisebox{\methodraise}{\textbf{#1}}%
  }%
}

\newcommand{\methodOne}[1]{%
  \methodBox{#1}%
}
\newcommand{\methodTwo}[2]{%
  \methodBox{\shortstack[c]{#1\\#2}}%
}

\newcommand{\cellMol}[2]{%
  \begin{minipage}[c]{\linewidth}\centering
    \includegraphics[width=\linewidth]{#1}\par\vspace{\imgformvspace}%
    \formula{#2}%
  \end{minipage}%
}

\newcommand{\cellMolCorrect}[2]{%
  \begingroup
  \setlength{\fboxrule}{1pt}
  \setlength{\fboxsep}{2pt}%
  \fcolorbox{CorrectGreen}{white}{%
    \begin{minipage}[c]{\linewidth}\centering
      \includegraphics[width=\linewidth]{#1}\par\vspace{\imgformvspace}%
      \formula{#2}%
    \end{minipage}%
  }%
  \endgroup
}

\theoremstyle{plain}

\theoremstyle{definition}

\theoremstyle{remark}

\title{Towards Reasonable Molecular Structure Elucidation \\ from Infrared Spectroscopy with Chemical Feedback}

\author{%
  Yusen Tan$^{1}$,
  Hongyu Zhan$^{1}$,
  Hai-tao Yu$^{1}$,
  Changxi Chi$^{2}$,
  Wenjie Du$^{3}$,
  Jun Xia$^{1,4,\dagger}$ \\
  $^{1}$The Hong Kong University of Science and Technology (Guangzhou) \\
  $^{2}$Westlake University \\
  $^{3}$University of Science and Technology of China \\
  $^{4}$The Hong Kong University of Science and Technology \\
  \texttt{\{ytan277, hzhan701, hyu382\}@connect.hkust-gz.edu.cn} \\
  \texttt{chichangxi@westlake.edu.cn} \\
  \texttt{duwenjie@mail.ustc.edu.cn} \\
  \texttt{junxia@hkust-gz.edu.cn}
}

\begin{document}

\maketitle

\begingroup
\renewcommand{\thefootnote}{\fnsymbol{footnote}}
\footnotetext[2]{Corresponding author.}
\endgroup

\begin{abstract}
Infrared (IR) spectra provide characteristic signals of molecular structure, which are often interpreted by experts via functional-group identification or library matching, making the process time-consuming and ambiguous. Recent machine learning methods have made progress in molecular structure elucidation using molecular formulas and IR spectra. However, these models often infer unreasonable candidate molecular structures, including top-ranked predictions. More specifically, the molecular formula implied by a candidate structure often fails to match the input molecular formula, and the candidate’s theoretical IR spectrum is often inconsistent with the observed IR spectrum. To address these issues, we propose \textbf{F}ormula- and \textbf{IR}-\textbf{M}atched \textbf{P}reference \textbf{O}ptimization (FIRMPO), a \textbf{general and plug-and-play} chemical feedback-driven preference optimization framework for molecular structure elucidation. FIRMPO incorporates chemical feedback as preference signals based on exact molecular formula matching and IR spectral consistency to guide reasonable structure predictions. Unlike generic preference optimization methods, FIRMPO is tailored to molecular structure elucidation while remaining \textbf{model-agnostic}, enabling it to be readily integrated with different structure prediction models in this class. This encourages models to prioritize structures that satisfy the chemical feedback, leading to a substantial improvement in the accuracy of top-ranked predictions. Extensive experiments on three widely used IR datasets show that FIRMPO significantly improves molecular structure elucidation accuracy over existing baselines.
\end{abstract}

\section{Introduction}

Analytical spectroscopic techniques, including Infrared (IR) spectroscopy, Mass Spectrometry (MS), and Nuclear Magnetic Resonance (NMR), are widely used in chemical analysis \cite{intro1, intro2}. By providing complementary information about molecular composition, functional groups, and atomic connectivity, they play a central role in molecular structure elucidation \cite{intro1,intro3}. Although IR spectroscopy typically provides less chemical detail than MS and NMR with respect to molecular weight, elemental composition, and stereochemistry, its low cost, minimal sample preparation, and rapid data acquisition make it a practical tool for preliminary characterization and high-throughput analysis \cite{intro4}. However, extracting complete structural information from IR spectra remains challenging due to complex spectral patterns \cite{intro5,intro6}. In most cases, interpretation of IR spectra still relies on expert-driven identification of characteristic functional groups or matching against reference libraries. This process is time-consuming and often ambiguous \cite{intro5,intro6,intro7}.

\begin{figure*}[!t]

\begin{center}
\centerline{\includegraphics[width=0.98\textwidth]{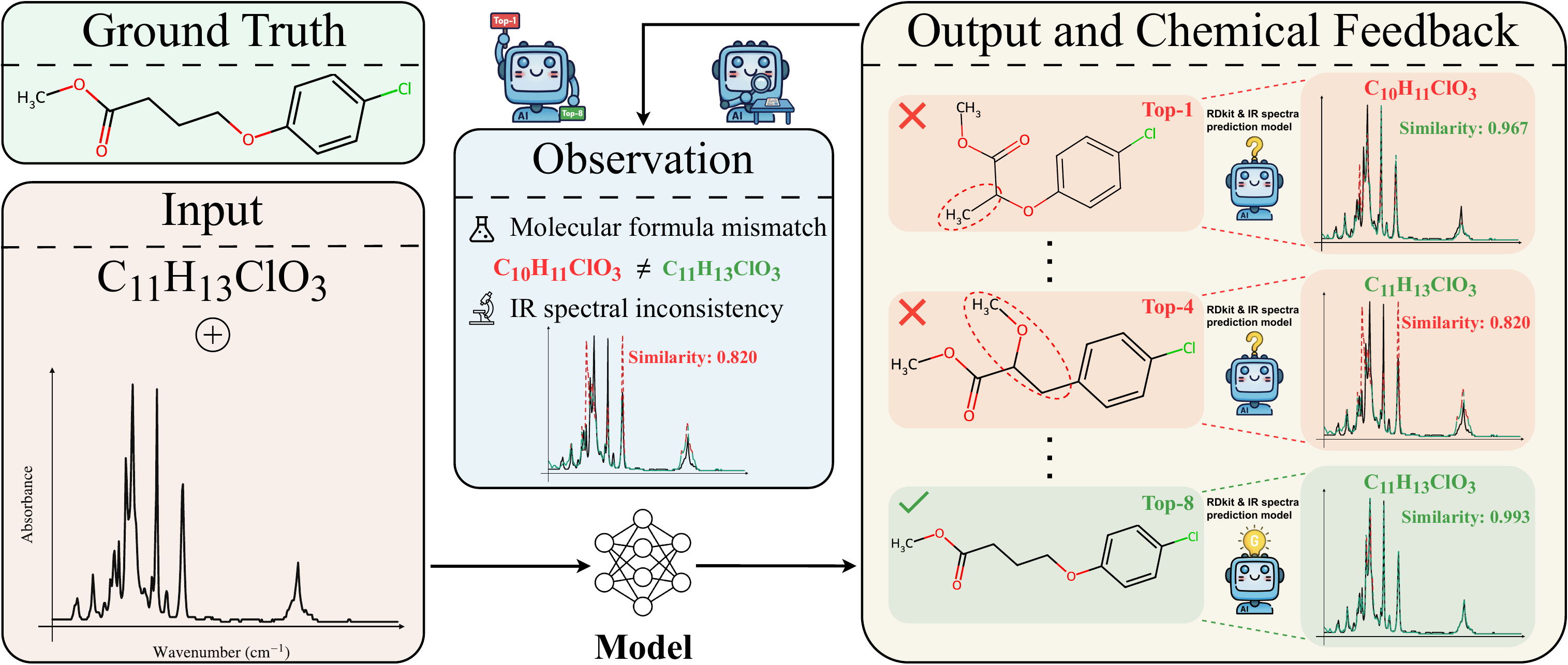}} 
\caption{Illustration of a ranking failure in molecular structure elucidation from molecular formulas and IR spectra: although the correct structure appears at rank 8, higher-ranked predictions violate chemical feedback, showing (i) molecular formula mismatch or (ii) inconsistency between the theoretical and observed IR spectra.}
\label{Fig-problem}
\end{center}

\end{figure*}

Recently, machine learning methods have been explored to directly infer molecular structures from molecular formulas and IR spectra. For example, IRtoMol employs a Transformer-based architecture conditioned on molecular formulas and IR spectra to enable end-to-end molecular structure elucidation \cite{intro15}. Subsequent studies have further advanced IR-based molecular structure prediction by introducing patch-based self-attention spectrum embeddings and improved Transformer architectures with refined spectral representations and decoding strategies\cite{patch_transformer, benchmark_ir}. Despite this progress, current models can still suffer from ranking failures: even when a plausible molecular structure is included among the generated candidates, an unreasonable candidate may be assigned the top-ranked position, as illustrated in Figure~\ref{Fig-problem}. This leads to a pronounced gap between top-$1$ and top-$k$ accuracy. Concretely, a top-ranked candidate may imply a molecular formula that differs from the input molecular formula, and its theoretical IR spectrum may be inconsistent with the observed spectrum.

To address these issues, we propose \textbf{F}ormula- and \textbf{IR}-\textbf{M}atched \textbf{P}reference \textbf{O}ptimization (FIRMPO), a \textbf{general and plug-and-play} chemical feedback-driven preference optimization framework for molecular structure elucidation. FIRMPO is \textbf{model-agnostic} and can be readily integrated with different IR-based molecular structure prediction models. The core idea is to incorporate chemical feedback, namely exact molecular formula matching and IR spectral consistency, as preference signals to re-rank candidate molecular structures into a chemically preferred order, thereby constructing a preference dataset. Building on this preference dataset, FIRMPO encourages the model to prioritize candidate structures that exactly match the input molecular formula and exhibit higher IR spectral consistency with the observed spectrum, leading to improved accuracy of top-ranked molecular structure predictions.

Overall, our main contributions are summarized as follows:
\begin{enumerate}[label=\arabic*)]
\item We introduce \textbf{F}ormula- and \textbf{IR}-\textbf{M}atched \textbf{P}reference \textbf{O}ptimization (FIRMPO), a \textbf{general and plug-and-play} chemical feedback-driven preference optimization framework for molecular structure elucidation. FIRMPO incorporates exact molecular formula matching and IR spectral consistency as preference signals to guide models toward chemically reasonable structure predictions.

\item FIRMPO is \textbf{model-agnostic} and can be readily integrated with different IR-based molecular structure prediction models. Unlike generic preference optimization methods, FIRMPO is tailored to molecular structure elucidation by introducing chemistry-aware preference construction and two additional weighting terms, which provide fine-grained optimization signals for ranking chemically preferred candidate structures.

\item Extensive experiments on three public IR spectral datasets show that FIRMPO significantly improves molecular structure elucidation accuracy over existing baselines under multiple evaluation metrics, especially for top-ranked predictions.
\end{enumerate}

\section{Related Work}
\subsection{Molecular Structure Elucidation from Infrared Spectroscopy}
IR spectroscopy is widely used for molecular characterization due to its sensitivity to functional groups and vibrational patterns. Traditional IR analysis often relies on expert interpretation or spectral database matching, which can limit throughput and may not generalize across diverse molecular systems \cite{intro7}. To mitigate these limitations, recent studies have applied machine learning methods to automate the interpretation of IR spectra \cite{intro8,intro9,intro10}. Early efforts mainly focused on functional group identification. These methods used convolutional neural networks, random forests, and multilayer perceptrons to extract discriminative spectral features for classification \cite{intro11,intro12}. While such models support preliminary analysis and compound screening, they provide only partial structural information and do not recover molecular connectivity or topology. This has motivated research on complete molecular structure elucidation directly from molecular formulas and IR spectra \cite{intro13}.

Complete molecular structure elucidation from molecular formulas and IR spectra is more challenging than functional group identification because it requires capturing richer chemical information \cite{intro14}. Recent work has explored end-to-end structural elucidation from IR spectra, often with additional molecular formula information. IRtoMol introduced a Transformer-based approach that conditions SMILES generation on molecular formulas and IR spectra for end-to-end molecular structure elucidation \cite{intro15}. Subsequent studies have advanced this paradigm by improving spectral representations with patch-based self-attention \cite{patch_transformer} and refining Transformer architectures with enhanced augmentation and decoding strategies \cite{benchmark_ir}.
Despite this progress, existing models can still assign high ranks to chemically unreasonable candidate structures. In particular, the molecular formula implied by a high-ranked candidate may not exactly match the provided molecular formula, and the candidate's theoretical IR spectrum may be inconsistent with the observed spectrum. Such ranking failures lead to a pronounced gap between top-1 and top-$k$ accuracy, suggesting that chemically plausible structures may be generated but are not consistently prioritized. This motivates us to construct preference signals from chemical feedback, namely molecular formula matching and IR spectral consistency, to guide models toward ranking chemically plausible candidate structures at the top positions.

\subsection{Preference Optimization (PO)}
Preference optimization (PO) is a paradigm for aligning model behavior with preference signals, including human judgments and domain-specific criteria. More broadly, recent alignment research increasingly emphasizes PO as an efficient alternative to reinforcement learning from human feedback, where learning signals come from comparative feedback such as preferences over multiple candidate outputs rather than explicit target texts \cite{ouyang2022training,DPO}. In practice, PO methods are often categorized into pointwise, pairwise, and listwise objectives, with recent work placing greater emphasis on the latter two \cite{DPO,KTO,LIPO,KPO}. Pairwise PO learns from comparisons between two candidates under the same context. For instance, Direct Preference Optimization (DPO) optimizes a closed-form, KL-regularized objective over preferred and dispreferred response pairs, enabling stable offline alignment without an explicit reward model \cite{DPO}. Kahneman--Tversky Optimization (KTO) instead uses pointwise good and bad labels inspired by prospect theory, reducing reliance on strictly constructed pairs \cite{KTO}.
While pairwise PO is simple and effective, it can underutilize supervision available in multi-candidate rankings. Listwise PO addresses this by learning from preferences defined over candidate sets, which naturally matches ranking tasks. Listwise Preference Optimization (LiPO) formulates alignment as learning to rank with listwise supervision over multiple candidates \cite{LIPO}.
K-order Ranking Preference Optimization (KPO) further emphasizes top-$K$ ranking consistency and introduces query-adaptive $K$ selection and curriculum learning \cite{KPO}. 

Despite their success in large language model alignment, existing PO methods are not readily applicable to molecular structure elucidation. In particular, these PO objectives do not explicitly prioritize candidates that are identical to the ground-truth structure, which is precisely the goal of molecular structure elucidation. Consequently, the correct structure can be ranked below a large candidate set of highly similar structural isomers, limiting improvements in molecular structure elucidation accuracy. This motivates us to develop a preference optimization method tailored to molecular structure elucidation that explicitly promotes ground-truth-matching candidates.

\subsection{Theoretical Infrared Spectra and Spectral Similarity Metrics}
A practical way to assess whether a candidate molecular structure is consistent with the observed IR spectrum is to predict its theoretical IR spectrum from the structure and then measure its similarity to the observed IR spectrum. Chemprop-IR~\cite{chemprop_ir} provides a learned forward model that predicts IR spectra from SMILES (Simplified Molecular Input Line Entry System) strings using message passing neural networks.

In this work, we quantify spectral consistency using cosine similarity between the predicted theoretical spectrum and the observed spectrum. Given vectorized spectra $\mathcal{X}_{\text{theory}}$ and $\mathcal{X}_{\text{obs}}$, cosine similarity is defined as:
\begin{equation}
\mathrm{CosSim}\!\left(\mathcal{X}_{\text{theory}}, \mathcal{X}_{\text{obs}}\right)
:=
\frac{\mathcal{X}_{\text{theory}}\cdot \mathcal{X}_{\text{obs}}}
{\left\lVert \mathcal{X}_{\text{theory}}\right\rVert_2 \left\lVert \mathcal{X}_{\text{obs}}\right\rVert_2}.
\end{equation}
Cosine similarity is scale-invariant and captures consistency in spectral shape, making it suitable for comparing theoretical and observed spectra.

\section{Methodology}
\subsection{Preliminaries}
Let $\mathcal{X}\in\mathbb{R}^{L}$ denote an IR spectrum discretized on $L$ wavenumber points, and let $\mathcal{F}$ denote the corresponding molecular formula.
Let $\mathcal{Y}$ denote the set of molecular structures represented by SMILES sequences.
For the task of molecular structure elucidation from molecular formulas and IR spectra, we denote the input by $x=(\mathcal{F},\mathcal{X})$ and the predicted output by $y\in\mathcal{Y}$, with the goal of generating $y$ identical to the ground-truth SMILES $y^\ast\in\mathcal{Y}$ given $x$.
However, this task is ill-posed in the sense that multiple distinct isomers, which are often structurally similar, can explain the same input $x$.
Therefore, in practice, the output is often represented as an ordered list of $M$ candidate structures $\mathbf{y}(x)=(y_1,\ldots,y_M)$, where $y_i\in\mathcal{Y}$.

Following \citet{intro15}, we formulate molecular structure elucidation from molecular formulas and IR spectra as learning a conditional generation policy:
\begin{equation}
\pi_\theta(\cdot\mid x): x \rightarrow \Delta(\mathcal{Y}),
\end{equation}
where $\Delta(\mathcal{Y})$ denotes the set of probability distributions over $\mathcal{Y}$. With the policy $\pi_\theta(\cdot\mid x)$, $\mathbf{y}(x)$ can be inferred via beam search given $x$.

\subsection{Supervised Learning for Molecular Structure Elucidation}
In prior work (e.g., IRtoMol~\citep{intro15}), $\pi_\theta(\cdot\mid x)$ is obtained via supervised learning on a dataset $\mathcal{D}=\{(x,y^\ast)\}$.
Representing the ground-truth SMILES as $y^\ast=(s_1,\ldots,s_T)$, $\theta$ is learned by minimizing the token-level negative log-likelihood:
\begin{equation}
\mathcal{L}_{\text{sup}}(\theta)=
\mathbb{E}_{(x,y^\ast)\sim \mathcal{D}}
\Bigl[
-\sum_{t=1}^{T}\log \pi_\theta\!\left(s_t \mid s_{<t}, x\right)
\Bigr].
\label{eq:sup_nll}
\end{equation}
This objective encourages $\pi_\theta$ to assign high probability to the ground-truth structure $y^\ast$, enabling it to perform molecular structure elucidation to a certain degree. However, the supervised policy $\pi_\theta$ often decodes chemically implausible candidate structures, as evidenced by violations of key chemical constraints such as exact molecular formula matching and IR spectral consistency.

\subsection{Preference Dataset Construction from Chemical Feedback}
To achieve more chemically plausible molecular structure elucidation, we employ preference optimization to update the supervised policy $\pi_\theta(\cdot\mid x)$, yielding an improved policy $\pi_{\theta'}(\cdot\mid x)$. A key prerequisite is a preference dataset that encodes chemically meaningful rankings over candidate structures. Accordingly, for each $x\in\mathcal{D}$, we first obtain an ordered list of candidate structures by decoding from the supervised policy $\pi_\theta(\cdot\mid x)$ via beam search. We then evaluate each candidate $y$ using chemical feedback from the input $x$, namely molecular formula matching and IR spectral consistency, and re-rank the candidates according to a chemistry-informed tiered lexicographic preference scheme. This produces a chemically preferred ordered list $\mathbf{y}^\ast(x)$, and collecting such lists over all inputs yields the desired preference dataset $\mathcal{D}^\ast=\{(x,\mathbf{y}^\ast(x))\}$. We next specify the chemical feedback metrics, the tiered preference scheme, and the construction procedure for $\mathcal{D}^\ast$.

\paragraph{Chemical feedback.}
We define chemical feedback as the combination of two chemically motivated metrics that can be computed from an input $x$ and a candidate structure $y$.
The first is whether the molecular formula matches, which can be expressed as:
\begin{equation}
m_{\hat{\mathcal{F}}}(y,\mathcal{F})
\;:=\;
\mathbb{I}\!\left[\hat{\mathcal{F}}(y)=\mathcal{F}\right]\in\{0,1\},
\end{equation}
where $\hat{\mathcal{F}}(y)$ denotes the molecular formula implied by the candidate structure $y$.
The second measures how consistent the theoretical IR spectrum is with the input spectrum, which can be formulated as:
\begin{equation}
s_{\mathrm{IR}}(y, \mathcal{X})
\;:=\;
\mathrm{CosSim}\!\bigl(g(y), \mathcal{X} \bigr),
\end{equation}
where $g(y)$ is a fixed predictor that maps a molecular structure $y$ to its theoretical IR spectrum.
In this work, we adopt Chemprop-IR~\cite{chemprop_ir} as $g(\cdot)$.

\paragraph{Chemistry-informed tiered preference scheme.}
Chemical feedback provides criteria for assessing the quality of candidate molecular structures.
Based on this feedback, we develop a tiered preference scheme with three priority levels:
(i) \textbf{highest priority $\mathcal{K}_1$}: candidates whose molecular structures are identical to the ground-truth molecular structure;
(ii) \textbf{intermediate priority $\mathcal{K}_2$}: among the remaining candidates, those whose molecular formulas exactly match the input molecular formula are preferred; within this formula-matched subset, candidates with higher IR spectral consistency are preferred;
(iii) \textbf{lowest priority $\mathcal{K}_3$}: among candidates that do not match the input molecular formula, those with higher IR spectral consistency are preferred.

Concretely, this tiered preference scheme can be implemented by assigning each candidate $y$ a lexicographic key, which can be expressed as:
\begin{equation}
k_x(y)
\;:=\;
\Bigl(
m_{\mathrm{Mol}}(y,y^\ast),\ 
m_{\hat{\mathcal{F}}}(y,\mathcal{F}),\ 
s_{\mathrm{IR}}(y,\mathcal{X})
\Bigr),
\label{eq:lex_key}
\end{equation}
where $m_{\mathrm{Mol}}(y,y^\ast)$ indicates whether the candidate molecular structure $y$ is identical to the ground-truth molecular structure $y^\ast$, which can be formulated as:
\begin{equation}
m_{\mathrm{Mol}}(y,y^\ast)
\;:=\;
\mathbb{I}\!\left[y=y^\ast\right]\in\{0,1\}.
\end{equation}

\paragraph{Construction of preference dataset $\mathcal{D}^\ast$.}
We define the preference dataset as $\mathcal{D}^\ast=\{(x,\mathbf{y}^\ast(x))\}$,
where each input $x$ is drawn from the supervised training dataset $\mathcal{D}$, and
$\mathbf{y}^\ast(x)=(y_1^\ast,\ldots,y_M^\ast)$ denotes a chemically preferred ordered list of $M$ candidate structures for $x$.
More specifically, for a given $x$, $\mathbf{y}^\ast(x)$ is constructed by sorting the candidate list $\mathbf{y}(x)$ in decreasing lexicographic order of $k_x(\cdot)$, which can be expressed as:
\begin{equation}
\mathbf{y}^\ast(x)
\;:=\;
\mathrm{sort\_descending}(\mathbf{y}(x), key=k_x).
\label{eq:def_Y_star}
\end{equation}
The complete procedure for constructing $\mathbf{y}^\ast(x)$ is provided in Appendix~\ref{app:algorithm}.

\begin{figure*}[!t]

\begin{center}
\centerline{\includegraphics[width=0.98\textwidth]{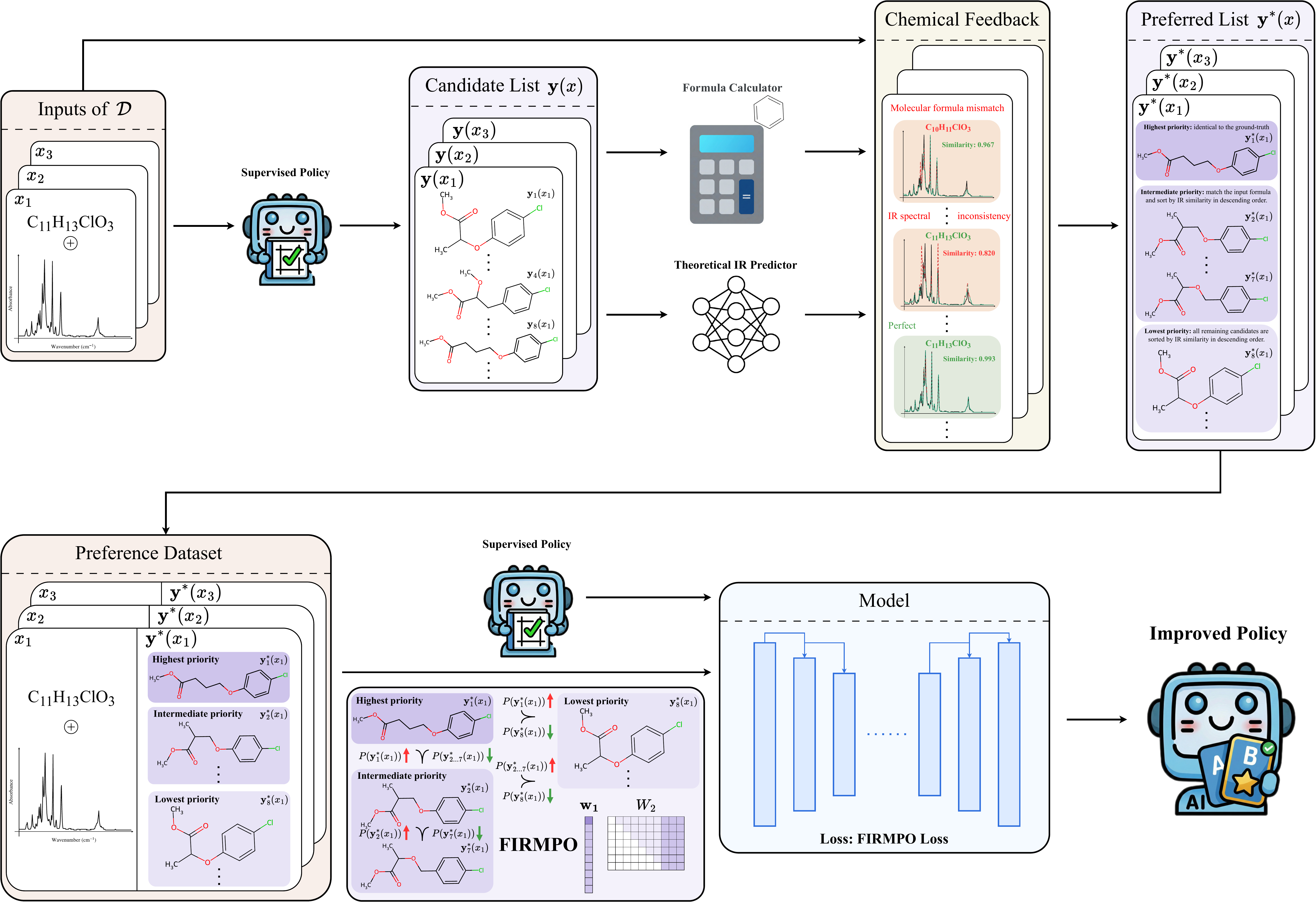}} 
\caption{Overview of the proposed method.}
\label{Fig-framwork}
\end{center}

\end{figure*}

\subsection{\textbf{F}ormula- and \textbf{IR}-\textbf{M}atched \textbf{P}reference \textbf{O}ptimization (FIRMPO) for Reasonable Molecular Structure Elucidation}
To more effectively exploit the preference signal encoded in $\mathcal{D}^\ast$ for policy updates and thereby promote reasonable molecular structure elucidation, we propose FIRMPO.
FIRMPO explicitly incorporates our chemistry-informed tiered preferences, i.e., exact molecular formula matching and IR spectral consistency, and its loss $\mathcal{L}^{\mathcal{K}(x)}_{\mathrm{FIRMPO}}$ is formulated as:
\begin{equation}
\begin{aligned}
& \mathcal{L}^{\mathcal{K}(x)}_{\mathrm{FIRMPO}}(\pi_{\theta'}; \pi_{\theta}) 
= - \mathbb{E}_{(x, y_1, \ldots, y_M) \sim \mathcal{D}^\ast} 
\Bigg[
\sum_{i=1}^{\mathcal{K}_2(x)} 
\log  \sigma  \Bigg(
- \log 
\sum_{j = \max(\mathcal{K}_1(x), i)+1}^{\mathcal{K}_3(x)} 
[\mathbf{w}_1]_i\exp \\ & \Bigg((
\beta[W_2]_{i,j}  \log \frac{\pi_{\theta'}(y_j|x)}{\pi_\theta(y_j|x)}
- \beta[W_2]_{i,j} \log \frac{\pi_{\theta'}(y_i|x)}{\pi_\theta(y_i|x)}
\Bigg)
\Bigg)
\Bigg],
\end{aligned}
\label{eq:firmpo_loss}
\end{equation}
where $\sigma(\cdot)$ is the sigmoid function and $\beta$ controls the strength of the regularization that keeps the updated policy $\pi_{\theta'}$ close to the reference policy $\pi_\theta$. $\mathcal{K}_1(x)$ denotes the number of candidates in the highest-priority tier $\mathcal{K}_1$, $\mathcal{K}_2(x)$ denotes the number of candidates in tiers $\mathcal{K}_1$ and $\mathcal{K}_2$, and $\mathcal{K}_3(x)$ denotes the total number of candidates in tiers $\mathcal{K}_1$, $\mathcal{K}_2$, and $\mathcal{K}_3$. The weight vector $\mathbf{w}_1$ and the pairwise strength coefficients $[W_2]_{i,j}$ (collected in a matrix $W_2$) are defined as follows:
\[
\mathbf{w}_1 =
\overbrace{[\,\mathcal{K}_2(x) \; \mathcal{K}_2(x) \; \cdots \; \mathcal{K}_2(x) \;}^{\mathcal{K}_1(x)}
\;\overbrace{\,1 \; \cdots \; 1\,]}^{\mathcal{K}_3(x)-\mathcal{K}_1(x)},
\]
\[
[W_2]_{i,j} :=
\begin{cases}
0, & j \le i,\\[2pt]
\frac{1}{\mathcal{K}_2(x)}, &  i< j \le \mathcal{K}_2(x),\\[4pt]
1, & \mathcal{K}_2(x) < j \le \mathcal{K}_3(x).
\end{cases}
\]
Here, the weight vector $\mathbf{w}_1$ upweights comparisons anchored at candidates in the highest-priority tier $\mathcal{K}_1$. The pairwise strength coefficients $[W_2]_{i,j}$ assign weaker strength to comparisons among top-ranked, formula-matched candidates ($i<j\le \mathcal{K}_2(x)$) and stronger strength to comparisons against lower-priority candidates ($\mathcal{K}_2(x)<j\le \mathcal{K}_3(x)$). The lower summation bound $\max(\mathcal{K}_1(x), i)+1$ in Eq.~\eqref{eq:firmpo_loss} avoids imposing an arbitrary within-tier ordering among $\mathcal{K}_1$ candidates.

Finally, the improved policy $\pi_{\theta'}(\cdot\mid x)$ is obtained by optimizing the FIRMPO objective:
\begin{equation}
\theta' \;=\; \arg\min_{\theta'}\ \mathcal{L}^{\mathcal{K}(x)}_{\mathrm{FIRMPO}}(\pi_{\theta'};\pi_\theta;\mathcal{D}^\ast).
\label{eq:po_problem_def}
\end{equation}
An overview of our method is illustrated in Fig.~\ref{Fig-framwork}.

\paragraph{Advantages of FIRMPO.}
Unlike other listwise preference optimization methods, FIRMPO is tailored to reasonable molecular structure elucidation by guiding the model to rank candidate molecular structures using chemical feedback. Compared with generic listwise methods, FIRMPO offers three key advantages:

\begin{enumerate}[label=\arabic*)]
    \item \textbf{Tier-aware optimization without imposing unnecessary within-tier order.}
    Instead of optimizing a fixed top-$K$ prefix that implicitly enforces a strict total order, FIRMPO respects our tiered preference structure. In particular, candidates in the top-priority tier $\mathcal{K}_1$ correspond to the same ground-truth molecule and therefore should be treated as equally preferred, rather than being assigned an arbitrary within-tier precedence.

    \item \textbf{Explicit upweighting for top-priority reference candidates.}
    FIRMPO explicitly upweights comparisons anchored at top-priority candidates ($y_i\in\mathcal{K}_1$) as reference points, thereby placing greater emphasis on the highest-priority tier and more effectively improving top-ranked predictions.
    
    \item \textbf{Non-uniform pairwise strength across tiers to emphasize decisive contrasts.}
    Rather than applying a uniform comparison strength to all pairs, FIRMPO assigns weaker strength to comparisons within the formula-matched region (tiers $\mathcal{K}_1$ and $\mathcal{K}_2$) and stronger strength to contrasts against lower-priority, formula-mismatched candidates (tier $\mathcal{K}_3$). This design encourages cross-tier separation dominate training while avoiding overly harsh constraints on subtle within-region distinctions.
\end{enumerate}

\section{Experiments}
\subsection{Experimental Setup}

\paragraph{Datasets.}
We evaluate on three IR spectrum--structure benchmarks covering both simulated and experimental spectra.
For simulated spectra, we use two datasets: (i) the QM9S synthetic IR dataset introduced by \citet{qm9s}, and (ii) the IBM synthetic IR dataset introduced by \citet{nipsdata}. These benchmarks provide paired IR spectra and molecular structures with known ground truth.
For experimental spectra, we use gas-phase IR spectra from the National Institute of Standards and Technology (NIST) Chemistry WebBook \citep{nistwebbook}, which reflects realistic measurement noise and artifacts absent from simulated data.
For all datasets, we randomly split the data into training, validation, and test sets with an 85/5/10\% ratio.

\paragraph{Baselines.}

We evaluate FIRMPO on three supervised IR-based structure elucidation backbones.
(i) \textbf{IRtoMol}~\citep{intro15} is a Transformer-based spectrum-to-SMILES model.
(ii) \textbf{PatchIR}~\citep{patch_transformer} improves spectral representation learning through patch-based self-attention.
(iii) \textbf{IR-Bench}~\citep{benchmark_ir} further strengthens Transformer-based IR modeling with improved spectral representations, data augmentation, and formula-constrained decoding.

For each backbone, we compare the original supervised model (\textbf{Base}) with its FIRMPO-optimized counterpart (\textbf{+FIRMPO}). This paired evaluation setup enables a direct assessment of whether FIRMPO consistently improves molecular structure elucidation performance across different architectures and datasets.

\paragraph{Evaluation metrics.}
We report top-$k$ molecular accuracy (\%), formula accuracy (\%), and IR similarity for $k\in\{1,5,10\}$.
All metrics are computed over the top-$k$ candidate set: an instance is counted as correct if any candidate among the top-$k$ satisfies the corresponding criterion.
Specifically, the top-$k$ molecular accuracy is
$\mathrm{Acc@}k = 100\cdot \mathbb{P}\!\left[\exists\, i\le k:\, m_{\mathrm{Mol}}(y_i,y^\star)=1\right]$,
where $\mathbb{P}[\cdot]$ denotes the empirical probability over the test set.
The top-$k$ formula accuracy is
$\mathrm{Acc}_{\mathrm{F}}@k = 100\cdot \mathbb{P}\!\left[\exists\, i\le k:\hat{\mathcal{F}}(y_i)=\mathcal{F}\right]$.
The top-$k$ IR similarity is
$\mathrm{Sim@}k = \mathbb{E}\!\left[\max_{1\le i\le k}\mathrm{Sim}\!\left(\mathcal{X},g(y_i)\right)\right]$,
where $\mathbb{E}[\cdot]$ denotes the average over the test set.

\begin{table*}[!t]
\centering
\caption{
Overall performance of supervised IR-based structure elucidation backbones and their FIRMPO-optimized variants on QM9S~\cite{qm9s}, IBM~\cite{nipsdata}, and NIST Chemistry WebBook~\cite{nistwebbook}. 
We report top-$k$ molecular accuracy (\%), formula accuracy (\%), and IR spectrum similarity. 
The best result for each metric within each dataset is highlighted in bold.
}
\label{tab:main_results}

\setlength{\tabcolsep}{3pt}
\renewcommand{\arraystretch}{1.05}

\resizebox{0.96\textwidth}{!}{%
\begin{tabular}{cllccccccccc}
\toprule

Dataset & Backbone & Variant
& \multicolumn{3}{c}{Molecule accuracy}
& \multicolumn{3}{c}{Formula accuracy}
& \multicolumn{3}{c}{IR spectrum similarity} \\

\cmidrule(lr){4-6}
\cmidrule(lr){7-9}
\cmidrule(lr){10-12}

& &
& Acc@1 & Acc@5 & Acc@10
& $\mathrm{Acc}_{\mathrm{F}}$@1 & $\mathrm{Acc}_{\mathrm{F}}$@5 & $\mathrm{Acc}_{\mathrm{F}}$@10
& Sim@1 & Sim@5 & Sim@10 \\

\midrule

\multirow{6}{*}{QM9S}

& \multirow{2}{*}{IRtoMol}
& Base
& 4.68 & 14.26 & 18.01
& 96.97 & 99.69 & 99.84
& 0.551 & 0.639 & 0.657 \\

& & +FIRMPO
& \textbf{6.48} & \textbf{16.53} & \textbf{19.96}
& \textbf{99.88} & \textbf{99.88} & \textbf{99.88}
& \textbf{0.662} & \textbf{0.663} & \textbf{0.665} \\

\cmidrule(lr){2-12}

& \multirow{2}{*}{PatchIR}
& Base
& 0.28 & 1.28 & \textbf{2.43}
& 99.84 & 99.94 & \textbf{99.95}
& 0.197 & 0.249 & \textbf{0.271} \\

& & +FIRMPO
& \textbf{0.29} & \textbf{1.29} & \textbf{2.43}
& \textbf{99.95} & \textbf{99.95} & \textbf{99.95}
& \textbf{0.270} & \textbf{0.271} & \textbf{0.271} \\

\cmidrule(lr){2-12}

& \multirow{2}{*}{IR-Bench}
& Base
& 2.35 & 12.32 & \textbf{22.65}
& 98.75 & 99.94 & \textbf{99.98}
& 0.199 & 0.251 & \textbf{0.270} \\

& & +FIRMPO
& \textbf{4.75} & \textbf{15.77} & \textbf{22.65}
& \textbf{99.98} & \textbf{99.98} & \textbf{99.98}
& \textbf{0.268} & \textbf{0.269} & \textbf{0.270} \\

\midrule

\multirow{6}{*}{IBM}

& \multirow{2}{*}{IRtoMol}
& Base
& 15.42 & 39.82 & 42.18
& 64.11 & 86.87 & 91.07
& 0.609 & 0.717 & 0.735 \\

& & +FIRMPO
& \textbf{25.45} & \textbf{41.12} & \textbf{44.97}
& \textbf{95.24} & \textbf{95.24} & \textbf{95.24}
& \textbf{0.719} & \textbf{0.747} & \textbf{0.754} \\

\cmidrule(lr){2-12}

& \multirow{2}{*}{PatchIR}
& Base
& 1.23 & 6.93 & \textbf{11.75}
& 75.63 & 95.65 & \textbf{98.28}
& 0.355 & 0.397 & \textbf{0.411} \\

& & +FIRMPO
& \textbf{4.31} & \textbf{9.81} & \textbf{11.75}
& \textbf{98.28} & \textbf{98.28} & \textbf{98.28}
& \textbf{0.402} & \textbf{0.407} & \textbf{0.411} \\

\cmidrule(lr){2-12}

& \multirow{2}{*}{IR-Bench}
& Base
& 69.48 & 86.10 & \textbf{88.90}
& 95.23 & 99.62 & \textbf{99.90}
& 0.367 & 0.392 & \textbf{0.405} \\

& & +FIRMPO
& \textbf{70.79} & \textbf{86.94} & \textbf{88.90}
& \textbf{99.90} & \textbf{99.90} & \textbf{99.90}
& \textbf{0.400} & \textbf{0.402} & \textbf{0.405} \\

\midrule

\multirow{6}{*}{\shortstack{NIST\\WebBook}}

& \multirow{2}{*}{IRtoMol}
& Base
& 36.51 & 54.33 & 56.68
& 67.33 & 82.05 & 85.15
& 0.728 & 0.856 & 0.877 \\

& & +FIRMPO
& \textbf{55.94} & \textbf{56.93} & \textbf{57.05}
& \textbf{86.88} & \textbf{86.88} & \textbf{86.88}
& \textbf{0.761} & \textbf{0.879} & \textbf{0.888} \\

\cmidrule(lr){2-12}

& \multirow{2}{*}{PatchIR}
& Base
& 15.35 & 37.25 & \textbf{47.28}
& 74.50 & 87.62 & \textbf{91.34}
& 0.783 & 0.886 & \textbf{0.910} \\

& & +FIRMPO
& \textbf{38.12} & \textbf{45.42} & \textbf{47.28}
& \textbf{91.34} & \textbf{91.34} & \textbf{91.34}
& \textbf{0.896} & \textbf{0.907} & \textbf{0.910} \\

\cmidrule(lr){2-12}

& \multirow{2}{*}{IR-Bench}
& Base
& 50.99 & 76.61 & \textbf{83.42}
& 96.04 & 98.64 & \textbf{98.89}
& 0.878 & 0.925 & \textbf{0.933} \\

& & +FIRMPO
& \textbf{62.38} & \textbf{79.58} & \textbf{83.42}
& \textbf{98.89} & \textbf{98.89} & \textbf{98.89}
& \textbf{0.917} & \textbf{0.928} & \textbf{0.933} \\

\bottomrule
\end{tabular}%
}
\end{table*}

\begin{figure*}[t]
  \centering

  \setlength{\aboverulesep}{0.5pt}
  \setlength{\belowrulesep}{0.5pt}
  \setlength{\tabcolsep}{3pt}
  \renewcommand{\arraystretch}{1.05}
\resizebox{0.95\textwidth}{!}{%
  \begin{tabular}{@{}
    >{\centering\arraybackslash}m{\methodcolw}
    >{\centering\arraybackslash}m{\topcolw}
    >{\centering\arraybackslash}m{\topcolw}
    >{\centering\arraybackslash}m{\topcolw}
    >{\centering\arraybackslash}m{\topcolw}
    >{\centering\arraybackslash}m{\topcolw}
  @{}}
    \toprule

    \multicolumn{1}{>{\columncolor{HeaderGray}\centering\arraybackslash}m{\methodcolw}}{\textbf{Model}} &
    \multicolumn{1}{>{\columncolor{HeaderGray}\centering\arraybackslash}m{\topcolw}}{\textbf{Top-1}} &
    \multicolumn{1}{>{\columncolor{HeaderGray}\centering\arraybackslash}m{\topcolw}}{\textbf{Top-2}} &
    \multicolumn{1}{>{\columncolor{HeaderGray}\centering\arraybackslash}m{\topcolw}}{\textbf{Top-3}} &
    \multicolumn{1}{>{\columncolor{HeaderGray}\centering\arraybackslash}m{\topcolw}}{\textbf{Top-4}} &
    \multicolumn{1}{>{\columncolor{HeaderGray}\centering\arraybackslash}m{\topcolw}}{\textbf{Top-5}} \\
    \midrule

    \addlinespace[\rowgap]

    \methodOne{IRtoMol} &
    \cellMol{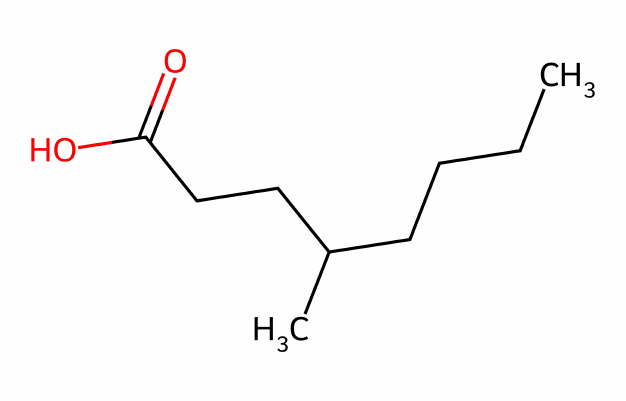}{} &
    \cellMol{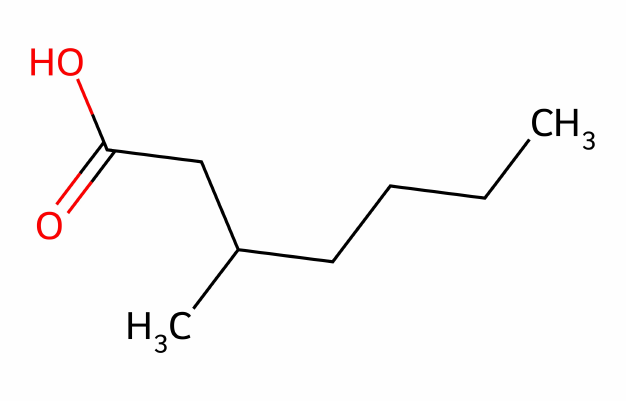}{} &
    \cellMolCorrect{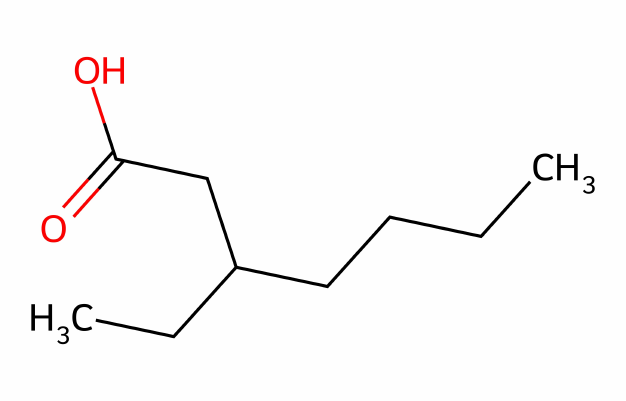}{} &
    \cellMol{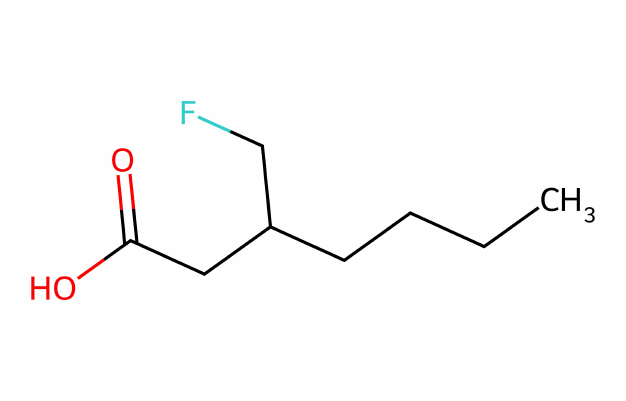}{} &
    \cellMol{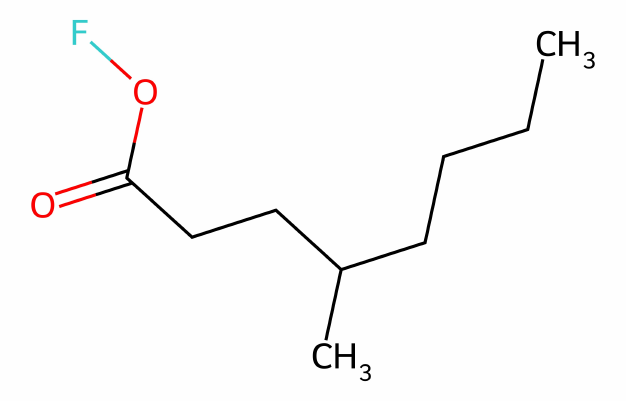}{} \\

    \addlinespace[\rowgap]

    \methodTwo{IRtoMol+}{FIRMPO} &
    \cellMolCorrect{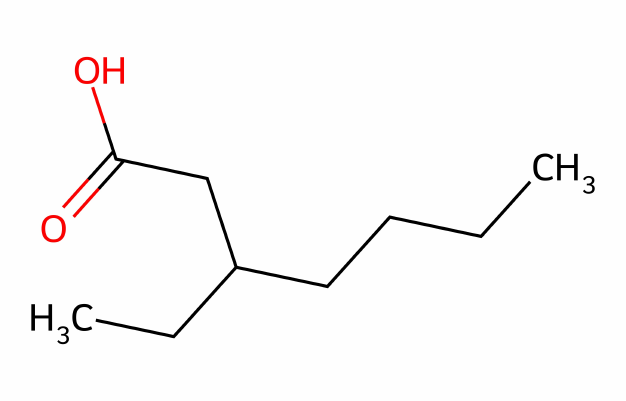}{} &
    \cellMol{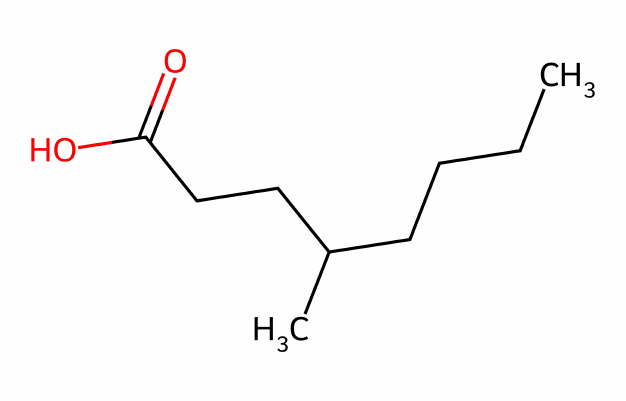}{} &
    \cellMol{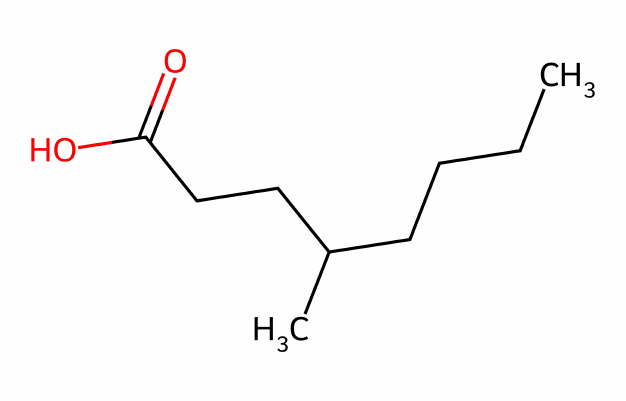}{} &
    \cellMol{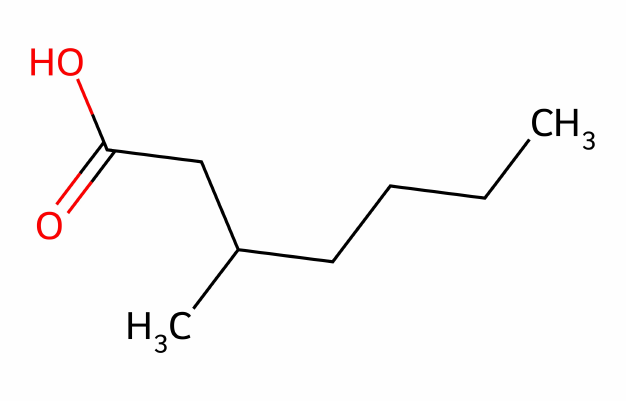}{} &
    \cellMol{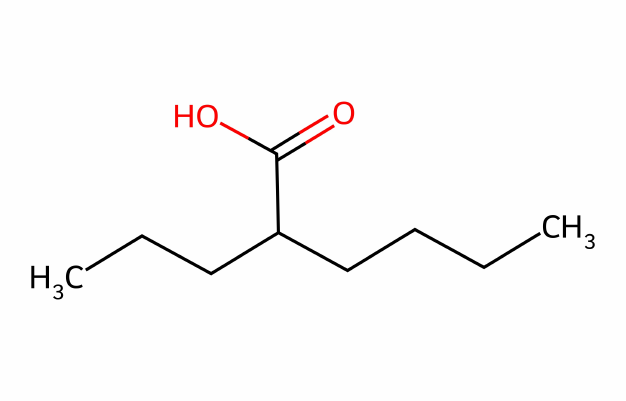}{} \\ [-1pt]

    \addlinespace[\rowgap]
    \bottomrule
  \end{tabular}
}
  \caption{Case study on a NIST Chemistry WebBook molecule (SMILES: \texttt{CCCCC(CC)CC(=O)O}) \citep{nistwebbook}, showing 2D depictions of the top-5 predicted candidate structures. Candidates matching the ground-truth structure are highlighted with green boxes.}
  \label{fig:qual_group_formula}
\end{figure*}

\subsection{Main Results}

Table~\ref{tab:main_results} reports overall performance on molecular structure elucidation from molecular formulas and IR spectra for QM9S~\cite{qm9s}, IBM~\cite{nipsdata}, and the NIST Chemistry WebBook~\cite{nistwebbook}. Across all three datasets and all evaluated backbones, applying FIRMPO consistently improves molecular accuracy, particularly at top-1. The improvements generally persist at top-5, indicating that FIRMPO enhances the overall ranking quality of generated candidates rather than improving only a single cutoff. The gains are accompanied by consistent improvements in both formula accuracy and IR spectrum similarity across different backbones and datasets, indicating that FIRMPO more effectively prioritizes chemically consistent candidates whose molecular formulas and theoretical IR spectra better match the input observations. Overall, the improvements in formula accuracy and IR spectrum similarity indicate that FIRMPO produces candidate structures that better satisfy chemical feedback, leading to more chemically reasonable molecular structure elucidation. More importantly, the consistent gains in molecular accuracy show that FIRMPO recovers the ground-truth structure more frequently by ranking chemically correct candidates higher.

\subsection{Case Study}

To qualitatively evaluate the effect of FIRMPO on molecular structure elucidation, we conduct a case study on a NIST WebBook molecule with SMILES \texttt{CCCCC(CC)CC(=O)O}. Figure~\ref{fig:qual_group_formula} shows the top-5 predicted candidate structures generated by IRtoMol and IRtoMol+FIRMPO. Across both models, most candidates preserve the carboxylic-acid scaffold, indicating that IR spectra strongly constrain the functional-group identity. The primary differences lie in the ranking of the ground-truth structure and the chemical plausibility of alternative candidates. 

\paragraph{IRtoMol.}
IRtoMol ranks two related carboxylic-acid isomers ahead of the ground-truth structure, placing the correct molecule at rank $3$. This behavior suggests that while the model successfully captures the dominant carboxylic-acid spectral signature, it struggles to distinguish subtle alkyl-chain branching patterns that produce highly similar IR spectra. Beyond the correct candidate, the model generates less plausible hypotheses, including fluorinated analogs, indicating relatively weak enforcement of molecular-formula consistency during decoding.

\paragraph{IRtoMol+FIRMPO (Ours).}
IRtoMol+FIRMPO places the ground-truth structure at rank $1$, demonstrating improved discrimination among closely related alkyl-branching isomers. Compared with the original IRtoMol model, the remaining candidates remain chemically plausible and formula-consistent, while avoiding spurious heteroatom substitutions.

\subsection{Ablations on Preference Signals}
\label{subsec:ablations_self}

We further examine the contribution of the preference signals used by FIRMPO to construct the chemically preferred ordered list \(\mathbf{y}^\ast(x)\). Since the main experiments evaluate FIRMPO across multiple supervised backbones, we use IRtoMol as a representative backbone in this analysis. This isolates the effect of different preference definitions while keeping the backbone architecture, candidate generation procedure, and preference optimization pipeline fixed.

\paragraph{Variants.}
All variants use the same IRtoMol backbone and the same candidate generation procedure, and differ only in how the ordered preference list is constructed during preference optimization. The IRtoMol baseline denotes the original supervised model without FIRMPO optimization, while IRtoMol+FIRMPO denotes the full model using the complete tiered preference definition in Eq.~\eqref{eq:lex_key}. We further consider two variants derived from the full model: (i) IRtoMol+FIRMPO w/o IR, which removes the IR-consistency preference signal, i.e., Eq.~\eqref{eq:lex_key} becomes \(k_x(y):=\bigl(m_{\mathrm{Mol}}(y,y^\star), m_{\hat{\mathcal{F}}}(y,\mathcal{F})\bigr)\); and (ii) IRtoMol+FIRMPO w/o Formula, which removes the formula-matching preference signal, i.e., Eq.~\eqref{eq:lex_key} becomes \(k_x(y):=\bigl(m_{\mathrm{Mol}}(y,y^\star), s_{\mathrm{IR}}(x,y)\bigr)\).

\paragraph{Results and interpretation.}
We conduct this analysis on the NIST Chemistry WebBook dataset. Table~\ref{tab:ablation_pref_signals} shows that all FIRMPO variants improve Acc@1 over the original IRtoMol baseline, indicating the effectiveness of preference optimization. Removing formula matching substantially reduces formula accuracy, while removing IR consistency preserves formula accuracy but weakens molecular ranking and IR similarity. The full IRtoMol+FIRMPO model achieves the best performance, and matches the best formula accuracy across all \(k\). These results show that formula matching and IR consistency provide complementary signals for chemistry-informed preference optimization.

\begin{table*}[t]
\centering
\caption{Ablation of preference signals on the NIST Chemistry WebBook dataset \cite{nistwebbook}, reporting top-1, top-5, and top-10 molecular accuracy (\%), formula accuracy (\%), and IR similarity. The best performance for each metric is in bold.}
\label{tab:ablation_pref_signals}

\setlength{\tabcolsep}{3pt}
\renewcommand{\arraystretch}{1.05}

\resizebox{0.9\textwidth}{!}{%
\begin{tabular}{lccccccccc}
\toprule
Model
& \multicolumn{3}{c}{Molecule accuracy}
& \multicolumn{3}{c}{Formula accuracy}
& \multicolumn{3}{c}{IR spectrum similarity} \\
\cline{2-10}
& Acc@1 & Acc@5 & Acc@10
& $\mathrm{Acc}_{\mathrm{F}}$@1 & $\mathrm{Acc}_{\mathrm{F}}$@5 & $\mathrm{Acc}_{\mathrm{F}}$@10
& Sim@1 & Sim@5 & Sim@10 \\
\midrule
IRtoMol
& 36.51 & 54.58 & 56.93
& 67.45 & 82.18 & 85.27
& 0.728 & 0.856 & 0.877 \\
IRtoMol+FIRMPO w/o IR
& 41.38 & 56.31 & \textbf{57.05}
& \textbf{86.88} & \textbf{86.88} & \textbf{86.88}
& 0.755 & 0.873 & 0.887 \\
IRtoMol+FIRMPO w/o Formula
& 52.97 & 56.56 & \textbf{57.05}
& 68.81 & 83.54 & 86.39
& 0.761 & 0.877 & 0.887 \\
IRtoMol+FIRMPO
& \textbf{55.94} & \textbf{56.93} & \textbf{57.05}
& \textbf{86.88} & \textbf{86.88} & \textbf{86.88}
& \textbf{0.762} & \textbf{0.879} & \textbf{0.888} \\
\bottomrule
\end{tabular}%
}
\end{table*}

\section{Conclusion}

This work narrows the gap between top-$1$ and top-$k$ molecular structure elucidation from molecular formulas and IR spectra. We propose \textbf{F}ormula- and \textbf{IR}-\textbf{M}atched \textbf{P}reference \textbf{O}ptimization (FIRMPO), a \textbf{general and plug-and-play} chemical feedback-driven preference optimization framework for molecular structure elucidation. FIRMPO is \textbf{model-agnostic} and integrates with IR-based structure prediction models. It constructs chemistry-informed preferences from exact molecular formula matching and IR spectral consistency, and optimizes models to prioritize chemically reasonable candidates that better satisfy these feedback signals. Experiments on two simulated datasets and one experimental IR dataset show that FIRMPO consistently improves molecular structure elucidation accuracy across different  backbone models, with gains persisting at top-$5$ and top-$10$.

\bibliography{example_paper}
\bibliographystyle{plainnat}

\newpage
\appendix

\section{Detailed Procedure for Constructing Preference Candidates}

The detailed procedure for constructing preference candidates \(\mathbf{y}^\ast(x)\) is provided in this section.

\label{app:algorithm}
\begin{algorithm}[h]
\caption{Chemical-feedback re-ranking with a tiered lexicographic key for constructing a preferred ordered list}
\label{alg:construct_Y_star}
\begin{algorithmic}[1]
\STATE {\bfseries Input:} $x=(\mathcal{F},\mathcal{X})$; ground-truth $y^\ast$; candidate list $\mathbf{y}(x)=(y_1,\ldots,y_M)$
\STATE {\bfseries Output:} Chemically preferred ordered list $\mathbf{y}^\ast(x)=(y_1^\ast,\ldots,y_M^\ast)$
\FOR{$i=1$ {\bfseries to} $M$}
    \IF{$y_i$ can be parsed into a valid molecule}
        \STATE $a_i \leftarrow \mathbb{I}\!\left[y_i=y^\ast\right]$
        \STATE $b_i \leftarrow \mathbb{I}\!\left[\hat{\mathcal{F}}(y_i)=\mathcal{F}\right] $
        \STATE $c_i \leftarrow \mathrm{CosSim}\!\bigl(g(y_i), \mathcal{X} \bigr)$
        \STATE $k_i \leftarrow (a_i,b_i,c_i)$
    \ELSE
        \STATE $k_i \leftarrow (-1,-1,-\infty)$
    \ENDIF
    
\ENDFOR
\STATE $\mathbf{y}^\ast(x) \leftarrow \mathrm{sort\_descending}(\mathbf{y}(x), key=k_x)$
\STATE \textbf{return} $\mathbf{y}^\ast(x)$
\end{algorithmic}
\end{algorithm}

\section{Experimental Details}
\label{app:exp_details}

All experiments are conducted on a single NVIDIA H100 GPU (80GB). Spectra are resampled over the range of 400--4000 cm$^{-1}$ and normalized using model-specific resolutions. For all three datasets, we use the same 85/5/10\% random split into training, validation, and test sets as in the main text. The validation set is used to select the best checkpoint and tune hyperparameters, including the learning rate, batch size, and preference-optimization strength $\beta$. For IRtoMol and IR-Bench, we initialize the models using the checkpoints released by the original authors and follow their recommended experimental setup.

\section{Additional Analysis of Preference Optimization}
\label{app:kpo_comparison}

In the main experiments, we evaluate FIRMPO by applying it to multiple supervised IR-based structure elucidation backbones and comparing each original supervised model with its FIRMPO-optimized counterpart. To complement these results, we include an additional analysis comparing our chemical feedback-driven preference optimization framework with a generic listwise preference optimization baseline.

Specifically, we compare \textbf{IRtoMol+FIRMPO} with \textbf{IRtoMol+KPO}. Both methods use the same supervised IRtoMol model as the initial backbone and are trained using the same preference data construction pipeline. The difference lies in the optimization objective: IRtoMol+KPO applies a generic listwise preference optimization objective, whereas IRtoMol+FIRMPO incorporates chemical feedback based on molecular formula matching and IR spectral consistency through the task-specific weighting terms \(\mathbf{w}_1\) and \(W_2\). These terms align the training objective with the tiered preference structure used in our task, which accounts for molecular correctness, formula consistency, and IR spectral similarity.

\begin{figure*}[t]
  \centering
  \begin{subfigure}[t]{0.3\linewidth}
    \centering
    \includegraphics[width=\linewidth]{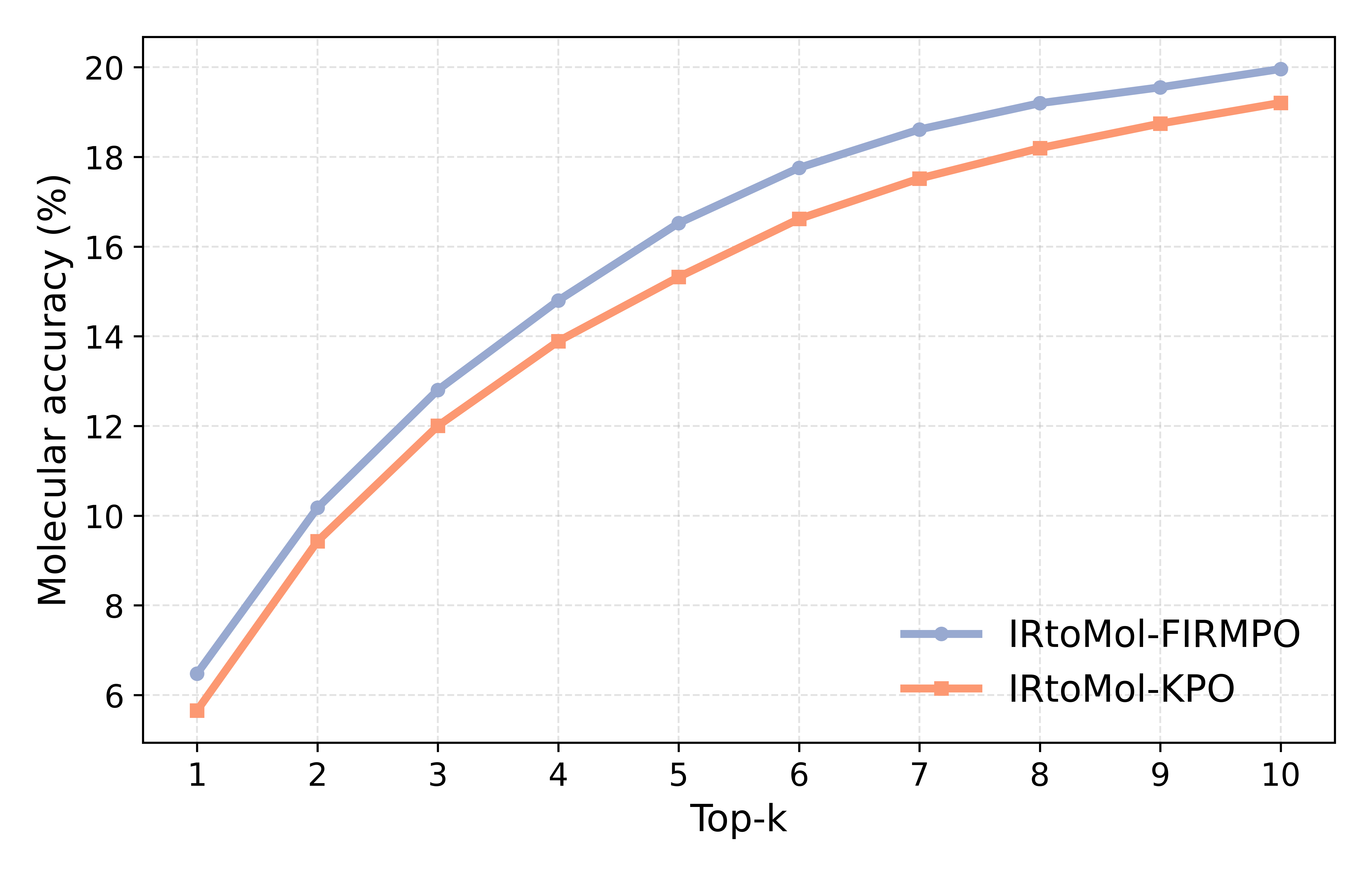}
    \caption{QM9S.}
    \label{fig:three_a}
  \end{subfigure}%
  \begin{subfigure}[t]{0.3\linewidth}
    \centering
    \includegraphics[width=\linewidth]{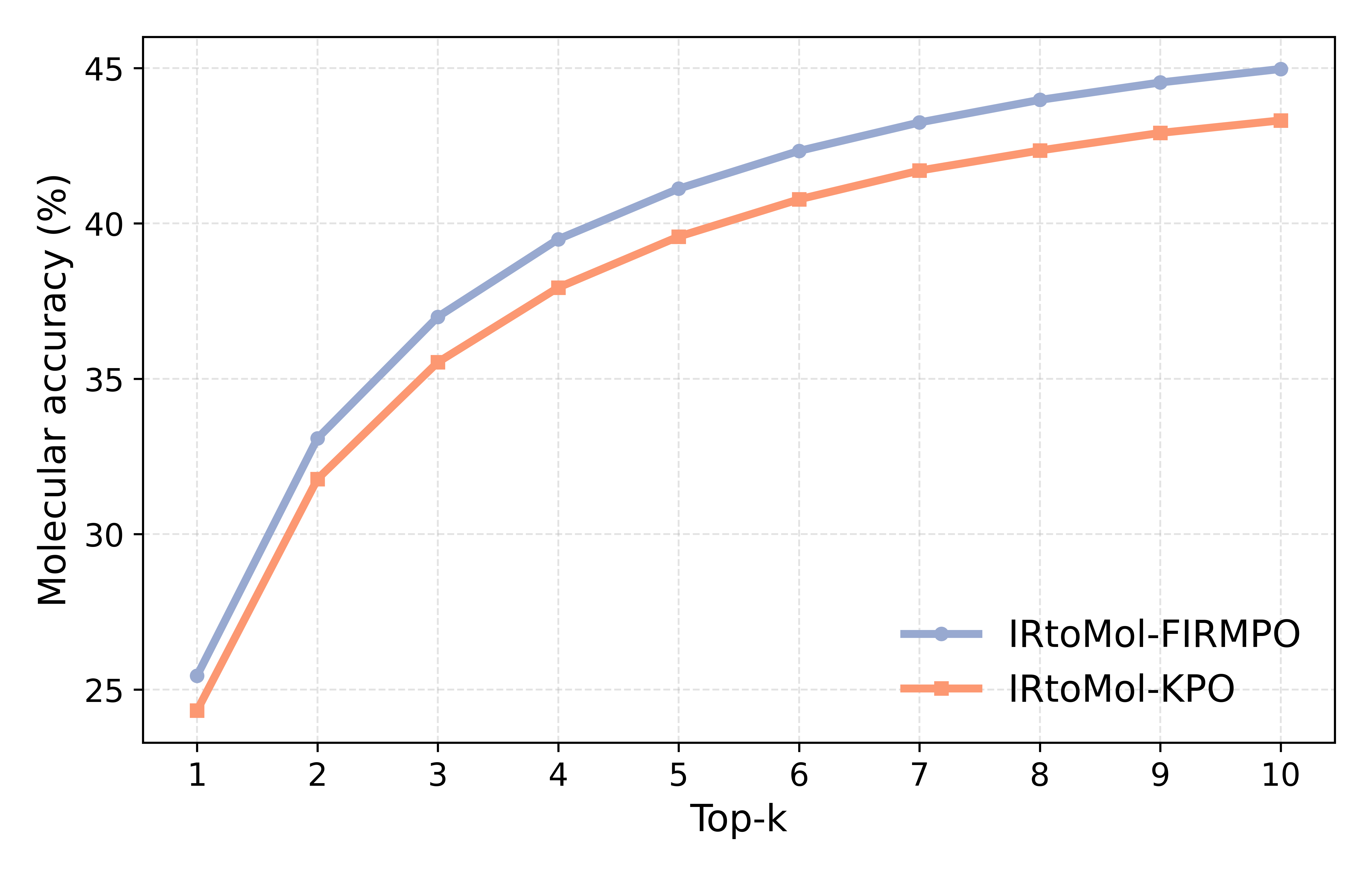}
    \caption{IBM.}
    \label{fig:three_b}
  \end{subfigure}%
  \begin{subfigure}[t]{0.3\linewidth}
    \centering
    \includegraphics[width=\linewidth]{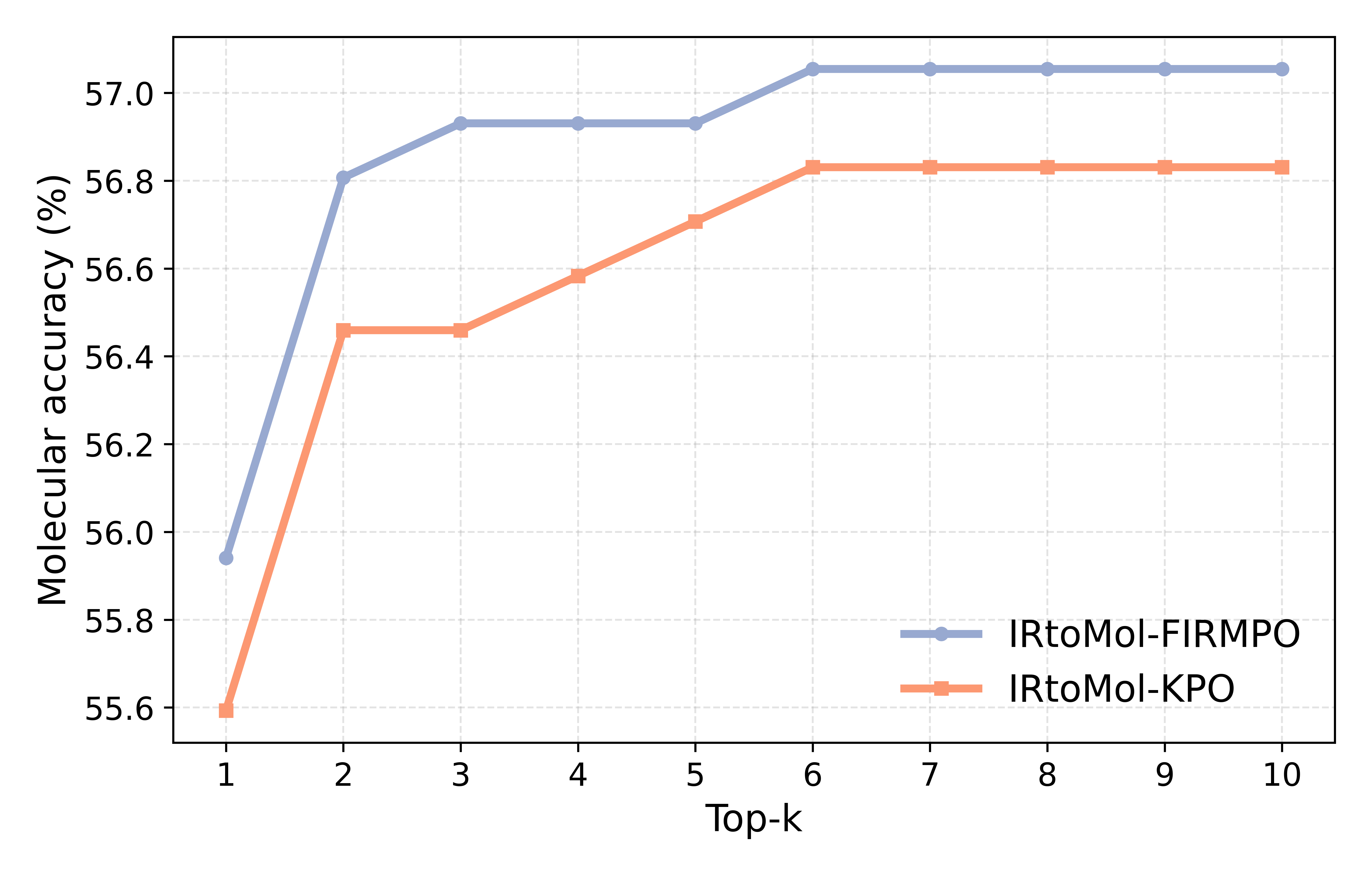}
    \caption{NIST Chemistry WebBook.}
    \label{fig:three_c}
  \end{subfigure}

  \caption{Top-$10$ molecular accuracy (\%) of IRtoMol+FIRMPO and IRtoMol+KPO on QM9S \cite{qm9s}, IBM \cite{nipsdata}, and NIST Chemistry Webbook \cite{nistwebbook}.}
  \label{fig:three}
\end{figure*}

Figure~\ref{fig:three} reports the top-\(k\) molecular accuracy for \(k \leq 10\) on QM9S, IBM, and the NIST Chemistry WebBook. Across all three datasets, IRtoMol+FIRMPO consistently achieves higher top-\(k\) molecular accuracy than IRtoMol+KPO for every evaluated value of \(k\). This result suggests that the improvement of FIRMPO is not merely due to the use of preference optimization in a generic form. Instead, the chemistry-informed weighting design provides additional benefits by better aligning the ranking objective with the evaluation criteria and the underlying structure of IR-based molecular prediction.

\section{Additional Case Studies}
\label{sec:appendix_case_studies}

To provide additional qualitative comparisons for the QM9S~\cite{qm9s} and IBM~\cite{nipsdata} datasets, this section presents case studies of molecular structure elucidation for several molecules from each dataset, focusing on the predictions generated by IRtoMol and IRtoMol+FIRMPO.

\begin{figure*}[h]
  \centering

  \setlength{\aboverulesep}{0.5pt}
  \setlength{\belowrulesep}{0.5pt}
  \setlength{\tabcolsep}{3pt}
  \renewcommand{\arraystretch}{1.05}
\resizebox{0.9\textwidth}{!}{%
  \begin{tabular}{@{}
    >{\centering\arraybackslash}m{\methodcolw}
    >{\centering\arraybackslash}m{\topcolw}
    >{\centering\arraybackslash}m{\topcolw}
    >{\centering\arraybackslash}m{\topcolw}
    >{\centering\arraybackslash}m{\topcolw}
    >{\centering\arraybackslash}m{\topcolw}
  @{}}
    \toprule

    \multicolumn{1}{>{\columncolor{HeaderGray}\centering\arraybackslash}m{\methodcolw}}{\textbf{Model}} &
    \multicolumn{1}{>{\columncolor{HeaderGray}\centering\arraybackslash}m{\topcolw}}{\textbf{Top-1}} &
    \multicolumn{1}{>{\columncolor{HeaderGray}\centering\arraybackslash}m{\topcolw}}{\textbf{Top-2}} &
    \multicolumn{1}{>{\columncolor{HeaderGray}\centering\arraybackslash}m{\topcolw}}{\textbf{Top-3}} &
    \multicolumn{1}{>{\columncolor{HeaderGray}\centering\arraybackslash}m{\topcolw}}{\textbf{Top-4}} &
    \multicolumn{1}{>{\columncolor{HeaderGray}\centering\arraybackslash}m{\topcolw}}{\textbf{Top-5}} \\
    \midrule

    \addlinespace[\rowgap]

    \methodOne{IRtoMol} &
    \cellMol{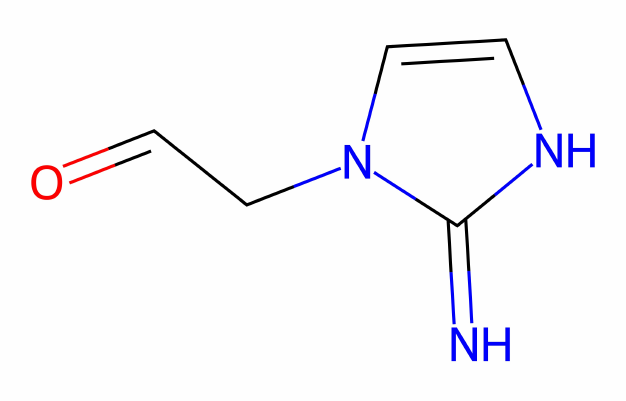}{} &
    \cellMol{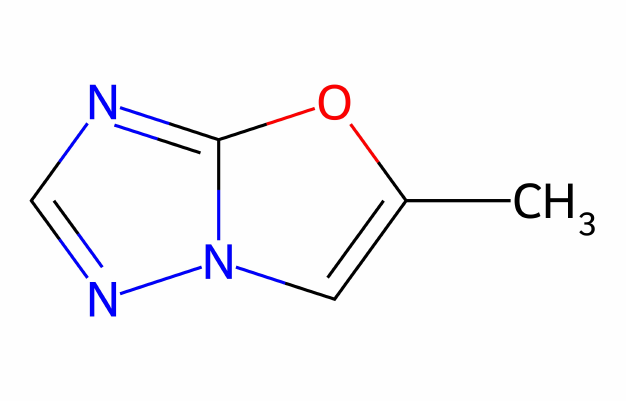}{} &
    \cellMolCorrect{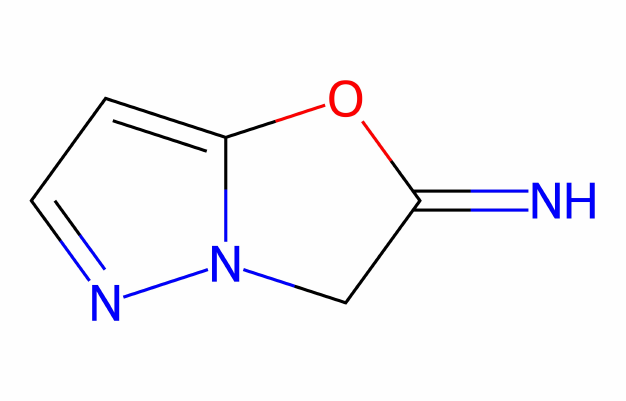}{} &
    \cellMol{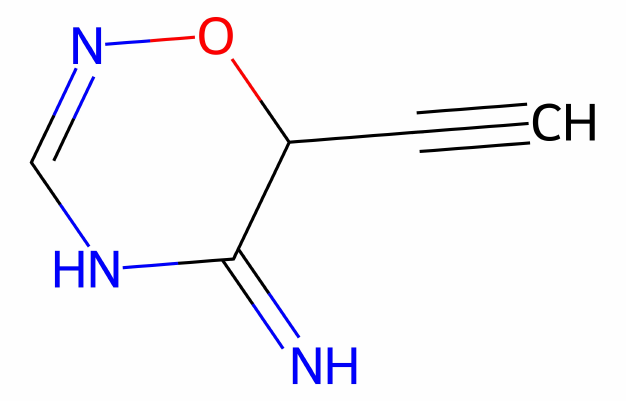}{} &
    \cellMol{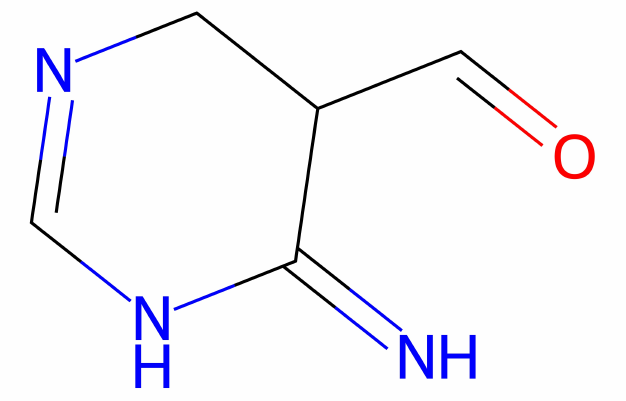}{} \\

    \addlinespace[\rowgap]

    \methodTwo{IRtoMol+}{FIRMPO} &
    \cellMolCorrect{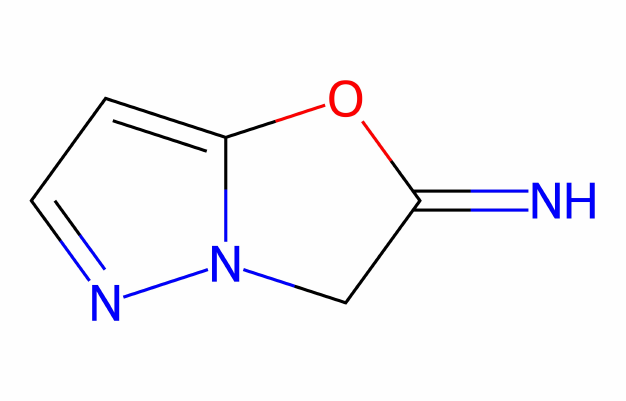}{} &
    \cellMol{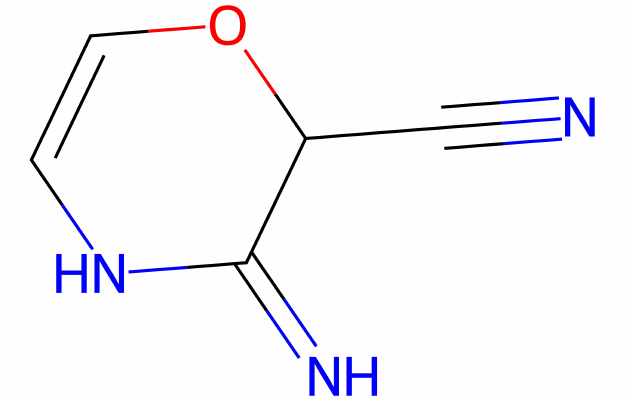}{} &
    \cellMol{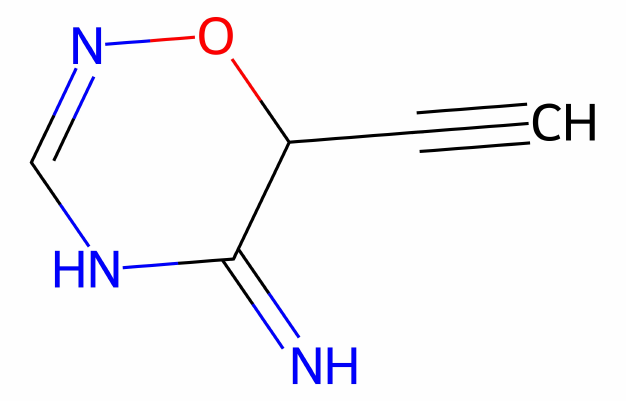}{} &
    \cellMol{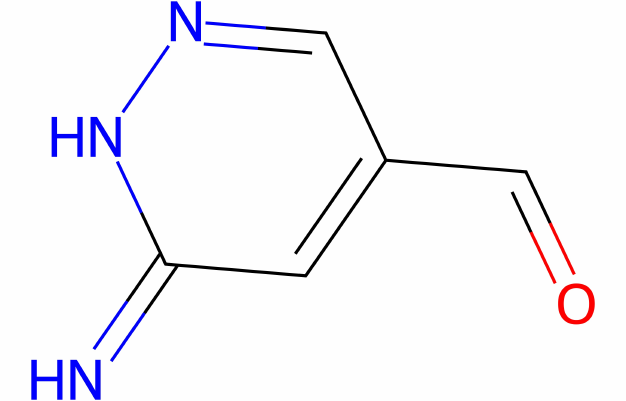}{} &
    \cellMol{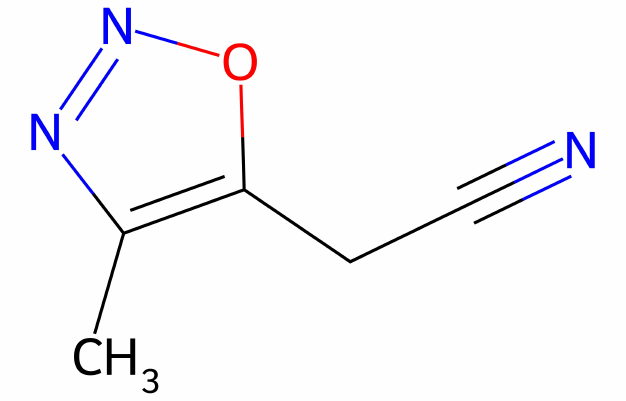}{} \\ [-1pt]

    \addlinespace[\rowgap]
    \bottomrule
  \end{tabular}
}
  \caption{Case study on a QM9S molecule (SMILES: \texttt{C\#CC1C2OC(=N)C12O}) \citep{qm9s}, showing 2D depictions of the top-5 predicted candidate structures.}
  \label{fig:qual_group_formula_qm9s}
\end{figure*}

\begin{figure*}[h]
  \centering

  \setlength{\aboverulesep}{0.5pt}
  \setlength{\belowrulesep}{0.5pt}
  \setlength{\tabcolsep}{3pt}
  \renewcommand{\arraystretch}{1.05}
\resizebox{0.9\textwidth}{!}{%
  \begin{tabular}{@{}
    >{\centering\arraybackslash}m{\methodcolw}
    >{\centering\arraybackslash}m{\topcolw}
    >{\centering\arraybackslash}m{\topcolw}
    >{\centering\arraybackslash}m{\topcolw}
    >{\centering\arraybackslash}m{\topcolw}
    >{\centering\arraybackslash}m{\topcolw}
  @{}}
    \toprule

    \multicolumn{1}{>{\columncolor{HeaderGray}\centering\arraybackslash}m{\methodcolw}}{\textbf{Model}} &
    \multicolumn{1}{>{\columncolor{HeaderGray}\centering\arraybackslash}m{\topcolw}}{\textbf{Top-1}} &
    \multicolumn{1}{>{\columncolor{HeaderGray}\centering\arraybackslash}m{\topcolw}}{\textbf{Top-2}} &
    \multicolumn{1}{>{\columncolor{HeaderGray}\centering\arraybackslash}m{\topcolw}}{\textbf{Top-3}} &
    \multicolumn{1}{>{\columncolor{HeaderGray}\centering\arraybackslash}m{\topcolw}}{\textbf{Top-4}} &
    \multicolumn{1}{>{\columncolor{HeaderGray}\centering\arraybackslash}m{\topcolw}}{\textbf{Top-5}} \\
    \midrule

    \addlinespace[\rowgap]

    \methodOne{IRtoMol} &
    \cellMol{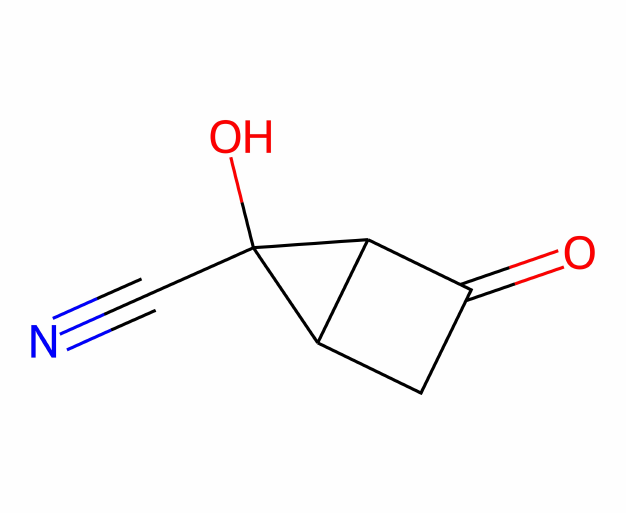}{} &
    \cellMol{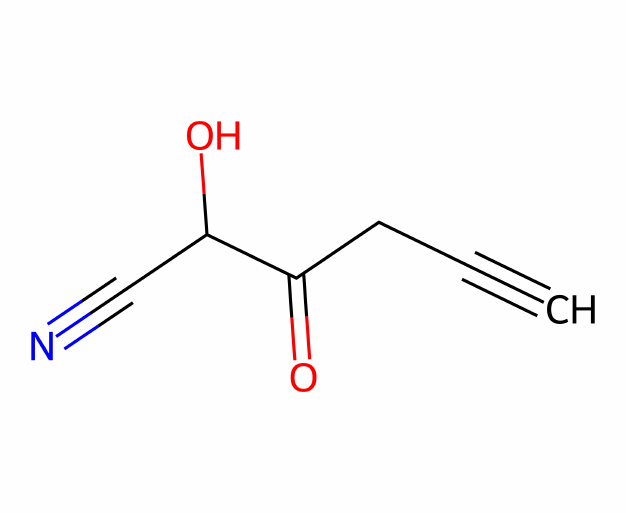}{} &
    \cellMol{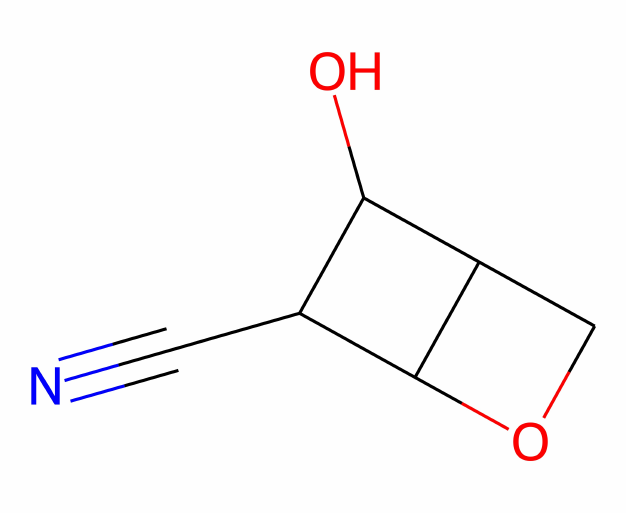}{} &
    \cellMol{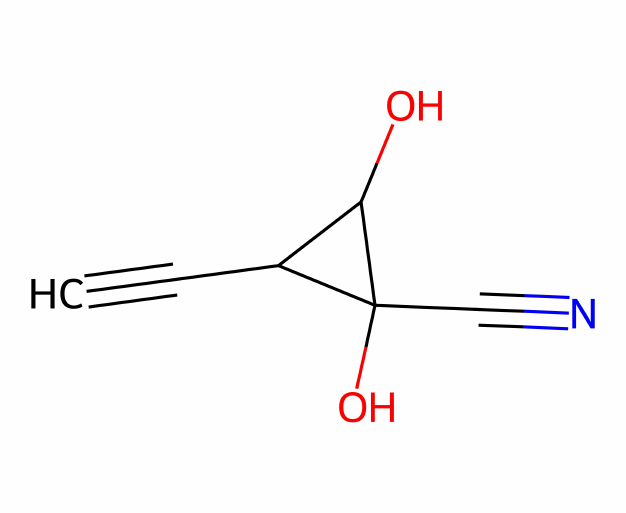}{} &
    \cellMol{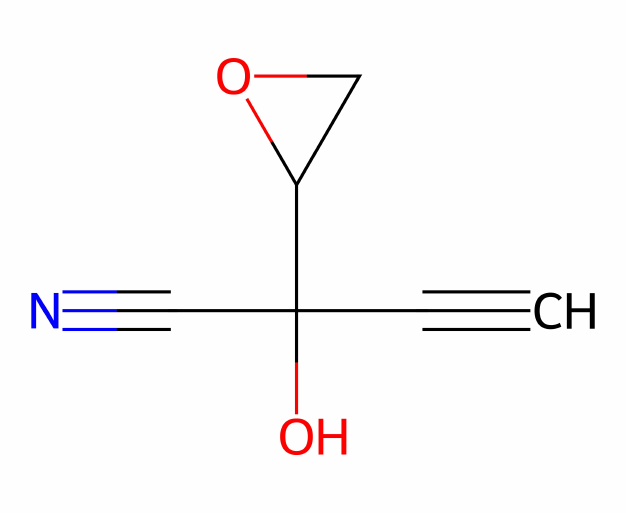}{} \\

    \addlinespace[\rowgap]

    \methodTwo{IRtoMol+}{FIRMPO} &
    \cellMol{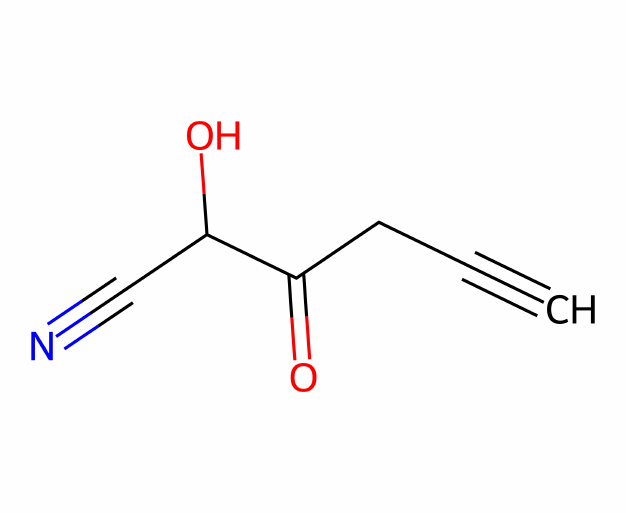}{} &
    \cellMolCorrect{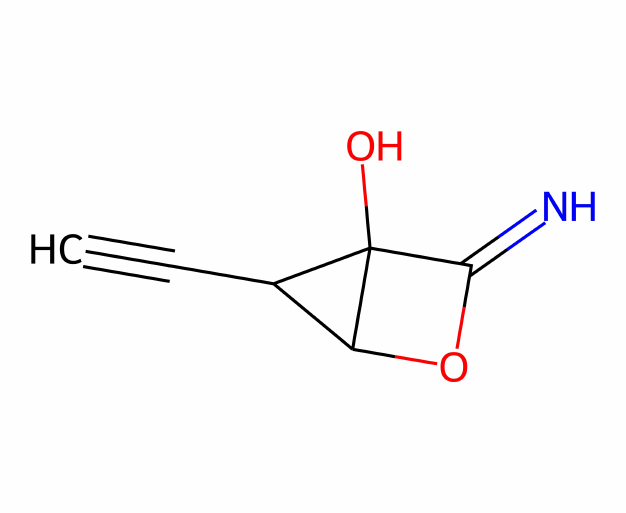}{} &
    \cellMol{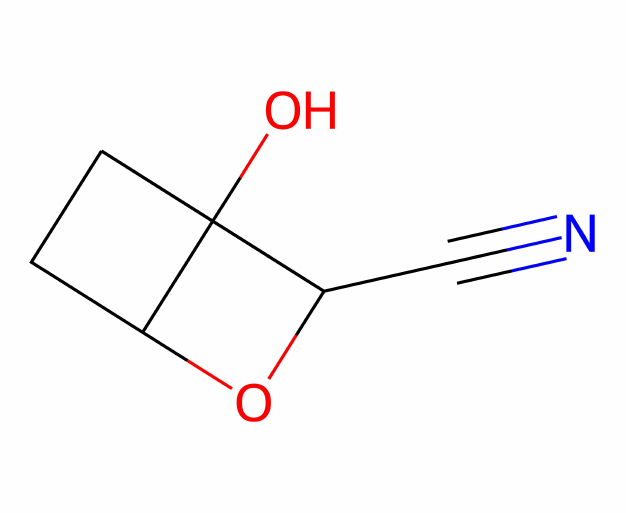}{} &
    \cellMol{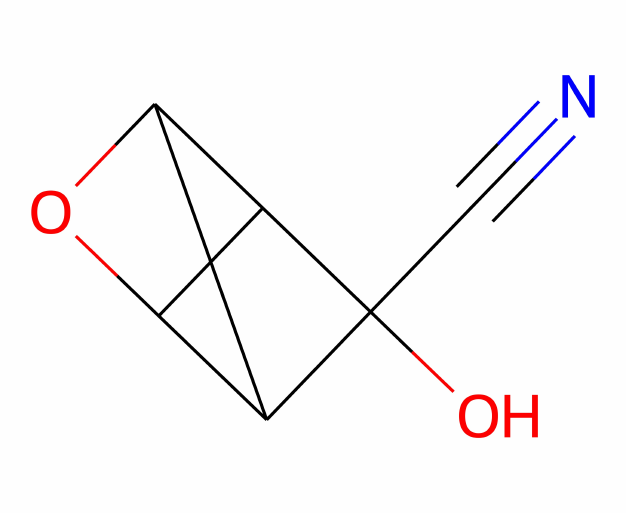}{} &
    \cellMol{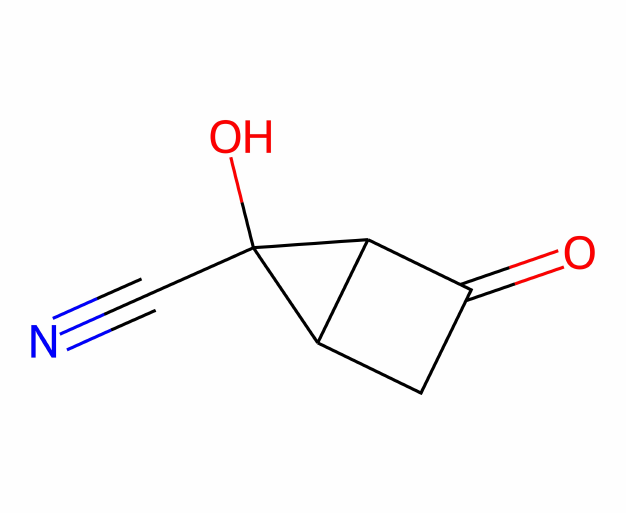}{} \\ [-1pt]

    \addlinespace[\rowgap]
    \bottomrule
  \end{tabular}
}
  \caption{Case study on a QM9S molecule (SMILES: \texttt{N=C1Cn2nccc2O1}) \citep{qm9s}, showing 2D depictions of the top-5 predicted candidate structures.}
  \label{fig:qual_group_formula_qm9s_1}
\end{figure*}

Figures~\ref{fig:qual_group_formula_qm9s} and \ref{fig:qual_group_formula_qm9s_1} show 2D depictions of the top-$5$ predicted candidate structures for two QM9S molecules with SMILES strings \texttt{C\#CC1C2OC(=N)C12O} and \texttt{N=C1Cn2nccc2O1}, respectively. First, for the molecule with SMILES \texttt{C\#CC1C2OC(=N)C12O}, IRtoMol ranks the ground-truth structure at rank $3$, while the remaining top predictions are closely related heterocyclic structural isomers. IRtoMol+FIRMPO ranks the ground-truth structure at rank $1$, and the remaining candidates remain chemically plausible and scaffold-consistent. Second, for the molecule with SMILES \texttt{N=C1Cn2nccc2O1}, IRtoMol does not recover the ground-truth structure within the top-$5$ and instead predicts strained small-ring isomers that preserve similar functional groups but alter the core connectivity. IRtoMol+FIRMPO ranks the ground-truth structure at rank $2$, and the remaining candidates remain chemically close while preserving the same functional-group inventory and overall ring topology.

\begin{figure*}[h]
  \centering

  \setlength{\aboverulesep}{0.5pt}
  \setlength{\belowrulesep}{0.5pt}
  \setlength{\tabcolsep}{3pt}
  \renewcommand{\arraystretch}{1.05}
\resizebox{0.9\textwidth}{!}{%
  \begin{tabular}{@{}
    >{\centering\arraybackslash}m{\methodcolw}
    >{\centering\arraybackslash}m{\topcolw}
    >{\centering\arraybackslash}m{\topcolw}
    >{\centering\arraybackslash}m{\topcolw}
    >{\centering\arraybackslash}m{\topcolw}
    >{\centering\arraybackslash}m{\topcolw}
  @{}}
    \toprule

    \multicolumn{1}{>{\columncolor{HeaderGray}\centering\arraybackslash}m{\methodcolw}}{\textbf{Model}} &
    \multicolumn{1}{>{\columncolor{HeaderGray}\centering\arraybackslash}m{\topcolw}}{\textbf{Top-1}} &
    \multicolumn{1}{>{\columncolor{HeaderGray}\centering\arraybackslash}m{\topcolw}}{\textbf{Top-2}} &
    \multicolumn{1}{>{\columncolor{HeaderGray}\centering\arraybackslash}m{\topcolw}}{\textbf{Top-3}} &
    \multicolumn{1}{>{\columncolor{HeaderGray}\centering\arraybackslash}m{\topcolw}}{\textbf{Top-4}} &
    \multicolumn{1}{>{\columncolor{HeaderGray}\centering\arraybackslash}m{\topcolw}}{\textbf{Top-5}} \\
    \midrule

    \addlinespace[\rowgap]

    \methodOne{IRtoMol} &
    \cellMol{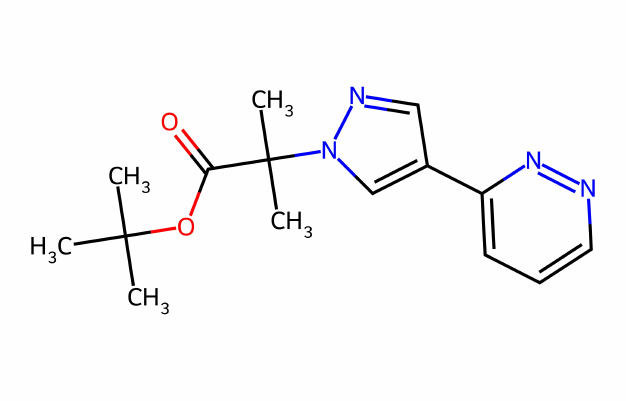}{} &
    \cellMol{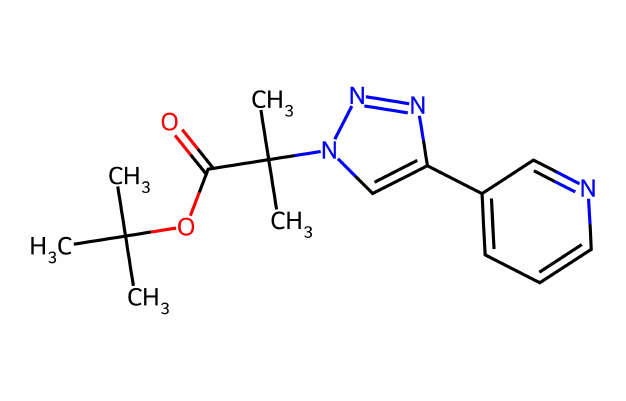}{} &
    \cellMol{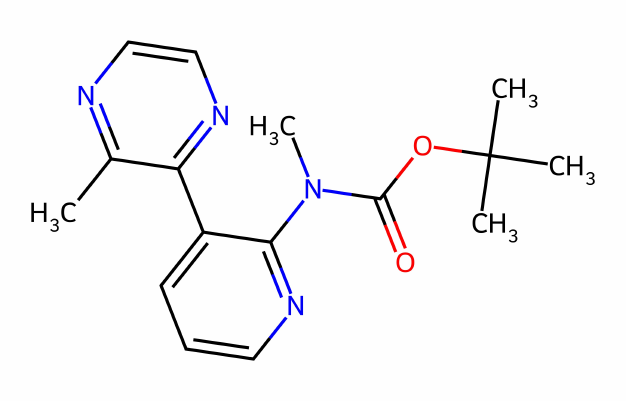}{} &
    \cellMolCorrect{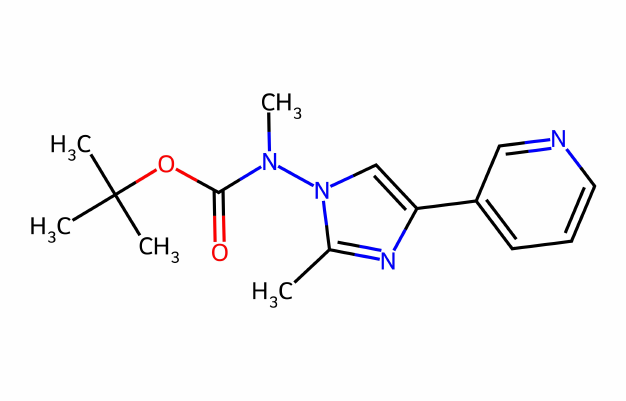}{} &
    \cellMol{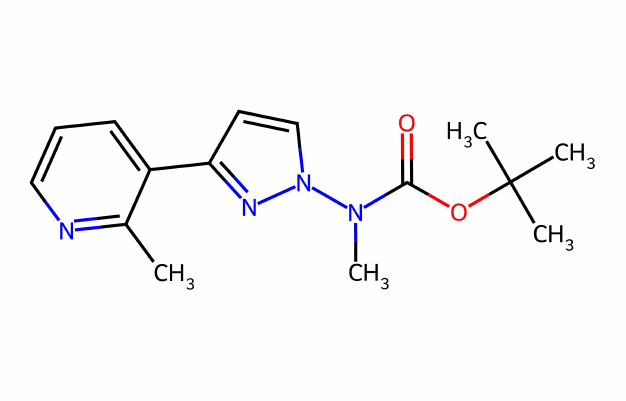}{} \\

    \addlinespace[\rowgap]

    \methodTwo{IRtoMol+}{FIRMPO} &
    \cellMolCorrect{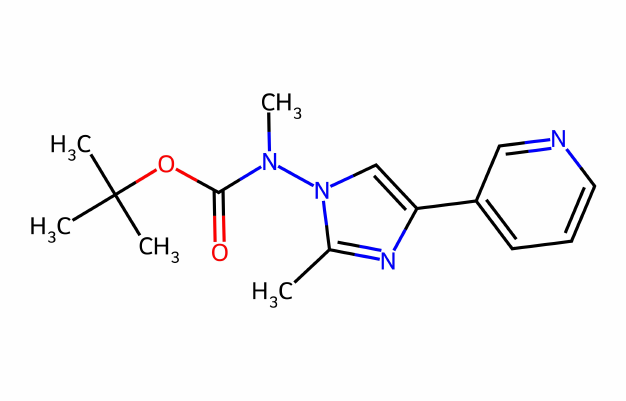}{} &
    \cellMol{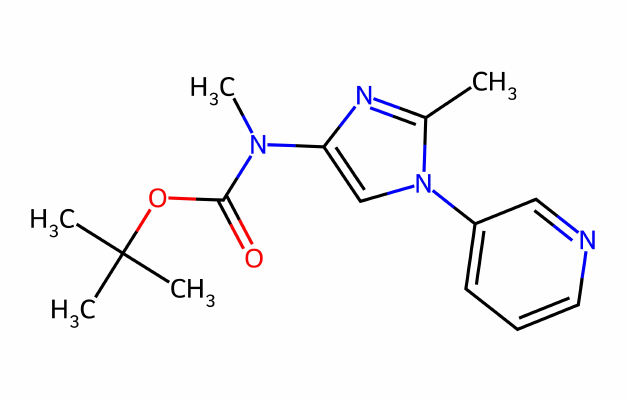}{} &
    \cellMol{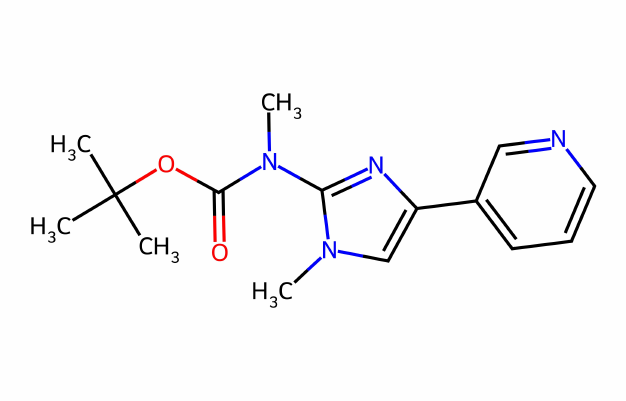}{} &
    \cellMol{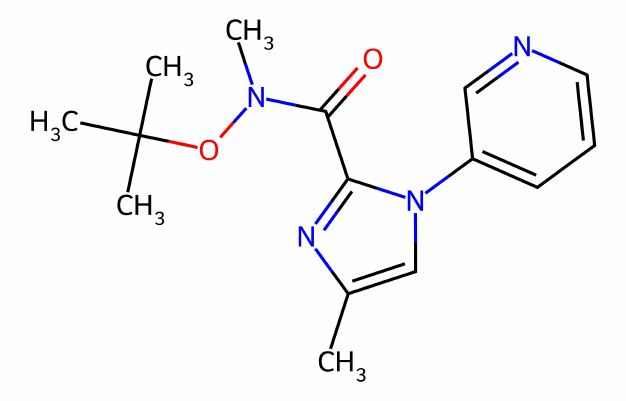}{} &
    \cellMol{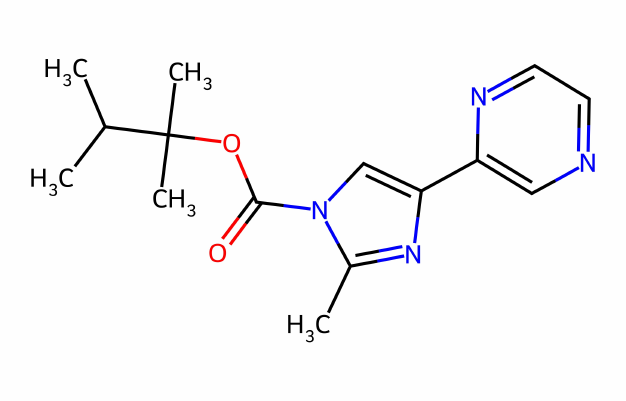}{} \\ [-1pt]

    \addlinespace[\rowgap]
    \bottomrule
  \end{tabular}
}
  \caption{Case study on a IBM molecule  (SMILES: \texttt{Cc1nc(-c2cccnc2)cn1N(C)C(=O)OC(C)(C)C}) \citep{nipsdata}, showing 2D depictions of the top-5 predicted candidate structures. }
  \label{fig:qual_group_formula_NIPS}
\end{figure*}

\begin{figure*}[h]
  \centering

  \setlength{\aboverulesep}{0.5pt}
  \setlength{\belowrulesep}{0.5pt}
  \setlength{\tabcolsep}{3pt}
  \renewcommand{\arraystretch}{1.05}
\resizebox{0.9\textwidth}{!}{%
  \begin{tabular}{@{}
    >{\centering\arraybackslash}m{\methodcolw}
    >{\centering\arraybackslash}m{\topcolw}
    >{\centering\arraybackslash}m{\topcolw}
    >{\centering\arraybackslash}m{\topcolw}
    >{\centering\arraybackslash}m{\topcolw}
    >{\centering\arraybackslash}m{\topcolw}
  @{}}
    \toprule

    \multicolumn{1}{>{\columncolor{HeaderGray}\centering\arraybackslash}m{\methodcolw}}{\textbf{Model}} &
    \multicolumn{1}{>{\columncolor{HeaderGray}\centering\arraybackslash}m{\topcolw}}{\textbf{Top-1}} &
    \multicolumn{1}{>{\columncolor{HeaderGray}\centering\arraybackslash}m{\topcolw}}{\textbf{Top-2}} &
    \multicolumn{1}{>{\columncolor{HeaderGray}\centering\arraybackslash}m{\topcolw}}{\textbf{Top-3}} &
    \multicolumn{1}{>{\columncolor{HeaderGray}\centering\arraybackslash}m{\topcolw}}{\textbf{Top-4}} &
    \multicolumn{1}{>{\columncolor{HeaderGray}\centering\arraybackslash}m{\topcolw}}{\textbf{Top-5}} \\
    \midrule

    \addlinespace[\rowgap]

    \methodOne{IRtoMol} &
    \cellMol{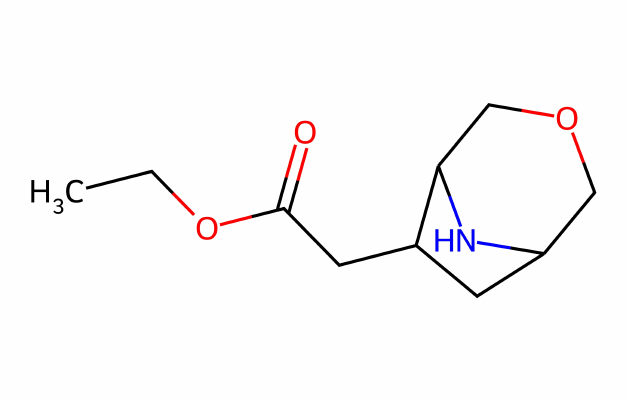}{} &
    \cellMol{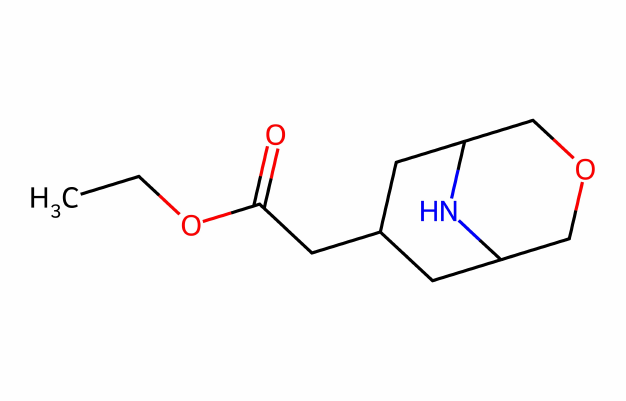}{} &
    \cellMol{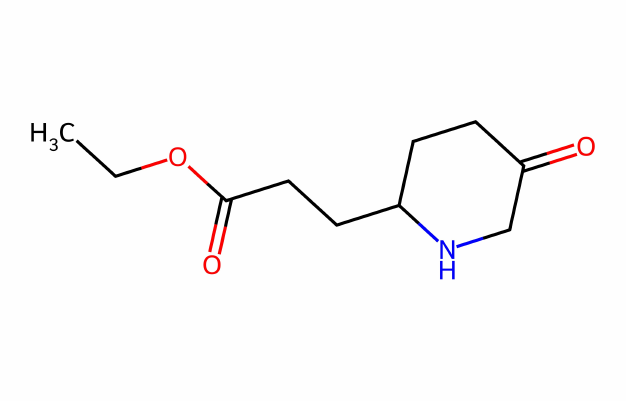}{} &
    \cellMol{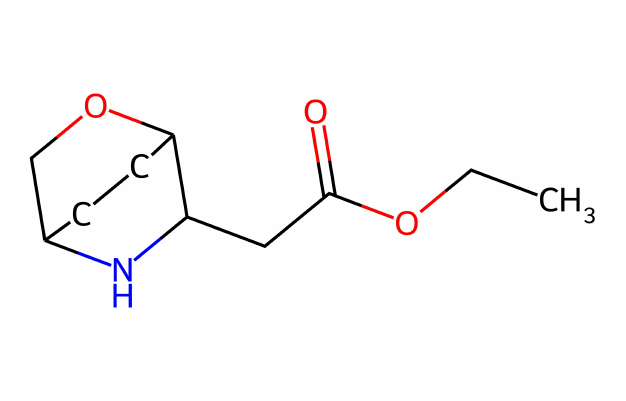}{} &
    \cellMol{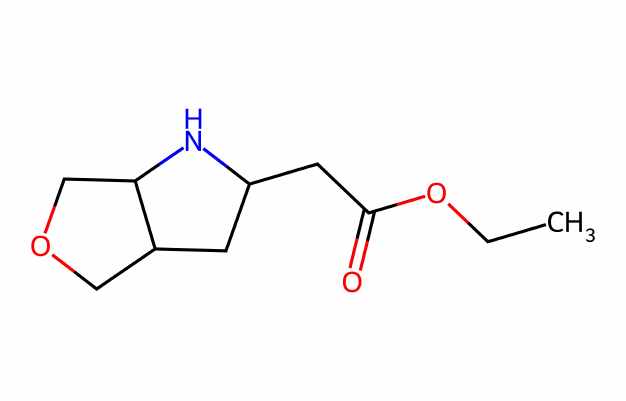}{} \\

    \addlinespace[\rowgap]

    \methodTwo{IRtoMol+}{FIRMPO} &
    \cellMol{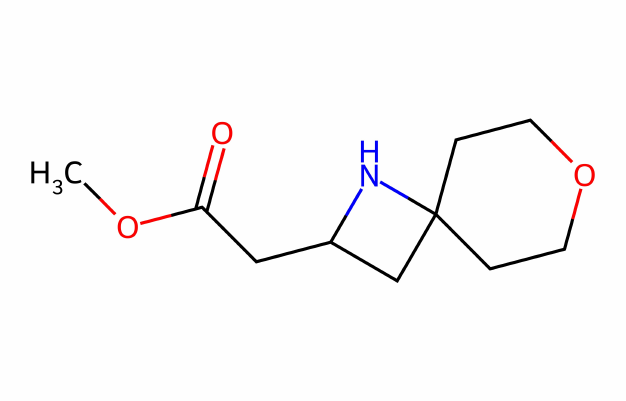}{} &
    \cellMolCorrect{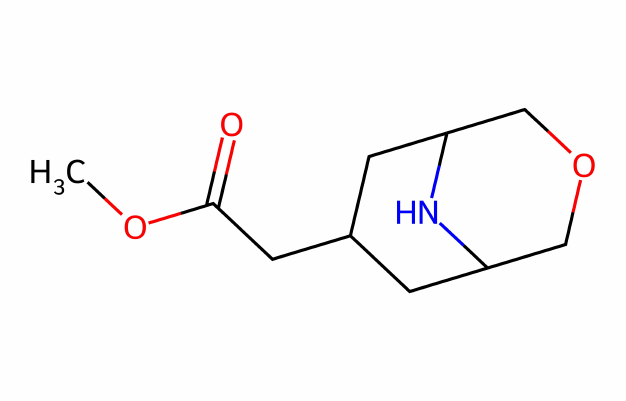}{} &
    \cellMol{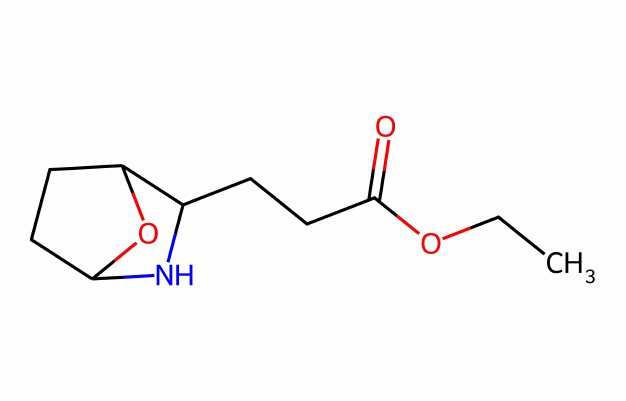}{} &
    \cellMol{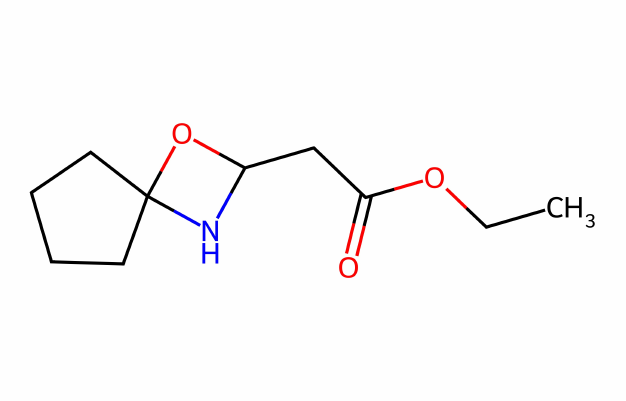}{} &
    \cellMol{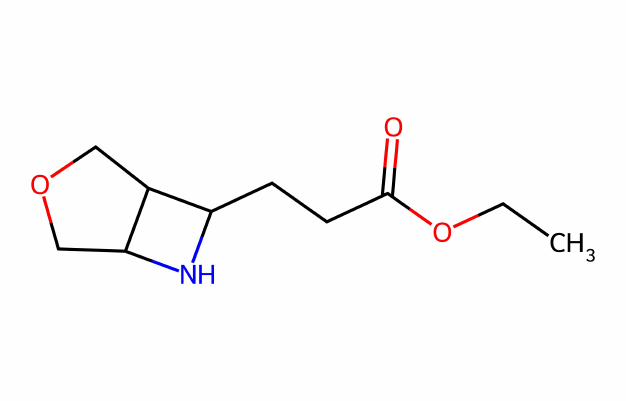}{} \\ [-1pt]

    \addlinespace[\rowgap]
    \bottomrule
  \end{tabular}
}
  \caption{Case study on a IBM molecule  (SMILES: \texttt{COC(=O)CC1CC2COCC(C1)N2}) \citep{nipsdata} , showing 2D depictions of the top-5 predicted candidate structures.}
  \label{fig:qual_group_formula_NIPS_1}
\end{figure*}

Figures~\ref{fig:qual_group_formula_NIPS} and \ref{fig:qual_group_formula_NIPS_1} show 2D depictions of the top-$5$ predicted candidate structures generated by IRtoMol and IRtoMol+FIRMPO for two IBM~\cite{nipsdata} molecules with SMILES strings \texttt{Cc1nc(-c2cccnc2)cn1N(C)C(=O)OC(C)(C)C} and \texttt{COC(=O)CC1CC2COCC(C1)N2}, respectively. First, for the molecule with SMILES \texttt{Cc1nc(-c2cccnc2)cn1N(C)C(=O)OC(C)(C)C}, IRtoMol ranks the ground-truth structure at rank $4$, while the remaining top-$5$ candidates are closely related regioisomers that preserve the same pyridyl imidazole carbamate scaffold but vary substituent placement and local connectivity. IRtoMol+FIRMPO ranks the ground-truth structure at rank $1$, and the remaining candidates remain chemically plausible and scaffold-consistent. Second, for the molecule with SMILES \texttt{COC(=O)CC1CC2COCC(C1)N2}, IRtoMol does not recover the ground-truth structure within the top-$5$ and instead predicts bicyclic amine and lactone-like variants with altered ring connectivity. IRtoMol+FIRMPO ranks the ground-truth structure at rank $2$, and the remaining candidates remain chemically close while preserving the same functional groups and overall bicyclic framework.

\section{Limitations and Future Directions}

Although FIRMPO consistently improves top-ranked molecular structure prediction, its performance still depends on the quality and diversity of the candidate structures generated by the underlying backbone model. If the correct structure is absent from the candidate set, the resulting preference supervision signals may be insufficient for effectively optimizing the model. Future work could investigate tighter integration between chemical feedback and candidate generation to further improve end-to-end molecular structure elucidation.


\end{document}